\documentclass[letterpaper]{article} 
\usepackage{aaai25}  
\usepackage{times}  
\usepackage{helvet}  
\usepackage{courier}  
\usepackage[hyphens]{url}  
\usepackage{graphicx} 
\usepackage{natbib}  
\usepackage{caption} 
\usepackage{amsmath,amssymb,amsfonts}
\usepackage{graphicx}
\usepackage{textcomp}
\usepackage{xcolor}
\def\BibTeX{{\rm B\kern-.05em{\sc i\kern-.025em b}\kern-.08em
    T\kern-.1667em\lower.7ex\hbox{E}\kern-.125emX}}

\usepackage{graphicx}
\usepackage{subfigure}
\usepackage[ruled,vlined,linesnumbered]{algorithm2e}
\usepackage{multirow}
\usepackage{booktabs}

\usepackage{bm}
\usepackage{amsfonts} 
\usepackage{stmaryrd} 

\usepackage{mathrsfs} 
\usepackage{todonotes}
\usepackage{comment}
 
\usepackage{color}
\definecolor{Orange}{rgb}{1,0.5,0}
\definecolor{Red}{rgb}{1,0,0}
\definecolor{Blue}{rgb}{0,0,1}

\usepackage{pifont}
\newcommand{\cmark}{\ding{51}}%
\newcommand{\xmark}{\ding{55}}%
\usepackage{newfloat}
\usepackage{listings}
\DeclareCaptionStyle{ruled}{labelfont=normalfont,labelsep=colon,strut=off} 
\title{Batch Selection for Multi-Label Classification Guided by Uncertainty and Dynamic Label Correlations}

\author{
    Ao Zhou\textsuperscript{\rm 1},
    Bin Liu \textsuperscript{\rm 1 \footnote{Corresponding author}},
    Jin Wang\textsuperscript{\rm 1},
    Grigorios Tsoumakas\textsuperscript{\rm 2}
}

\affiliations{
    \textsuperscript{\rm 1} Key Laboratory of Data Engineering and Visual Computing, School of Computer Science and Technology, Chongqing University of Posts and Telecommunications, Chongqing, China\\

   \textsuperscript{\rm 2} School of Informatics, Aristotle University of Thessaloniki, Thessaloniki, Greece\\
    zacqupt@gmail.com, \{liubin, wangjin\}@cqupt.edu.cn, greg@csd.auth.gr
}
\date{} 

\begin{document}
\maketitle

\begin{abstract}
The accuracy of deep neural networks is significantly influenced by the effectiveness of mini-batch construction during training. In single-label scenarios, such as binary and multi-class classification tasks, it has been demonstrated that batch selection algorithms preferring samples with higher uncertainty achieve better performance than difficulty-based methods. 
Although there are two batch selection methods tailored for multi-label data, none of them leverage important uncertainty information. Adapting the concept of uncertainty to multi-label data is not a trivial task, since there are two issues that should be tackled. First, traditional variance or entropy-based uncertainty measures ignore fluctuations of predictions within sliding windows and the importance of the current model state.
Second, existing multi-label methods do not explicitly exploit the label correlations, particularly the uncertainty-based label correlations that evolve during the training process.
In this paper, we propose an uncertainty-based multi-label batch selection algorithm. It assesses uncertainty for each label by considering differences between successive predictions and the confidence of current outputs, and further leverages dynamic uncertainty-based label correlations to emphasize instances whose uncertainty is synergistically expressed across multiple labels. 
Empirical studies demonstrate the effectiveness of our method in improving the performance and accelerating the convergence of various multi-label deep learning models.
\end{abstract}


%

\section{Introduction}

Multi-label classification (MLC) involves learning from instances associated with multiple labels simultaneously. Its goal is to derive a model capable of assigning a relevant set of labels to unseen instances.
For example, a news document might cover various topics in text categorization \cite{text1, text2}; an image could contain annotations for different scenes \cite{image1,image2}, and a video may consist of multiple different clips \cite{video1, video2}. For classifying such complex scenarios, multi-label learning approaches are seen as viable solutions for handling data with multiple labels.

Deep learning has recently proven successful in learning from multi-label data \cite{mlltrend}. 
By forming appropriate latent embedding spaces, deep neural networks manage to unravel the complex dependencies between features and labels in multi-label data~\cite{C2AE,MPVAE}. 
Moreover, deep learning models can successfully dissect and analyze label correlations \cite{CLIF,HOTVAE}. In addition, their inherent strength in representation learning allows them to naturally model label-specific features \cite{PACA}.

Recent studies highlight the critical role of mini-batch sample selection in the performance of deep neural networks (DNNs). Training with simple examples \cite{easy} can enhance robustness against outliers and noisy labels, but their smaller loss and gradients lead to slower model convergence. Conversely, focusing on instances that are difficult to predict correctly \cite{hardsample,hardsample2,OHEM} accelerates training, but overemphasizing the losses of hard examples may lead to overfitting on noisy data.
Uncertainty-based batch selection, exemplified by \textit{Active Bias} \cite{activebias}, is a compromise solution that prioritizes uncertain samples—those with unstable predictions, whether correct or incorrect, during the training process—thereby expediting model convergence while mitigating the risk of overfitting. \textit{Recency Bias} \cite{RecencyBias} evaluates instance uncertainty within recent observations rather than the entire history, offering a more dynamic assessment.

While batch selection methods are well-studied in single-label tasks (binary or multi-class classification), their effectiveness in multi-label data is less explored. 
\textit{Balanced} \cite{mllbalance} is a multi-label batch selection method that maintains label distributions in each batch consistent with the whole dataset via a re-weighting strategy. However, it only relies on the prior label distribution, neglecting losses or predictions of instances within each step of the training procedure. 
\textit{Hard Imbalance} \cite{mllour} prioritizes hard samples associated with more highly imbalanced (low-frequency) labels during multi-label batch selection, but it also suffers the risk of overfitting due to overemphasizing difficult instances. 

\begin{figure*}[!h]
    \centering
    \includegraphics[width=0.8\linewidth]{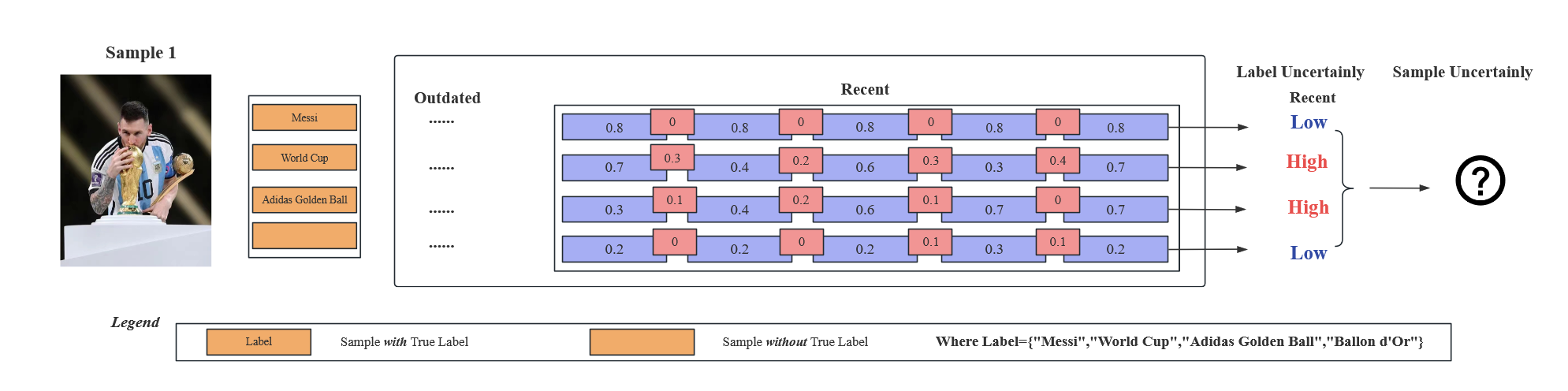}
    \caption{Sample 1 containing "Messi," "World Cup," and "Adidas Golden Ball," }
    \label{fig:intro}
\end{figure*}

This paper adapts the uncertainty-based batch selection strategy to multi-label data.
To achieve this, there are two issues that need to be addressed.
First, traditional variance and entropy-based measures \cite{activebias, RecencyBias} for assessing label-wise uncertainty ignore fluctuations within sliding windows, i.e., the changes between successive predictions, and the confidence of the current prediction. As the example shown in Fig \ref{fig:intro}. \textit{Active Bias} and \textit{Recent Bias} fail to distinguish the uncertainty of the labels "World Cup" and "Adidas Golden Ball" \footnote{Please refer to section 3.1 for an explanation of the reasons}. 
Secondly, directly summing all label uncertainties overlooks the correlation between labels. These inter-label uncertainties contain valuable information, reflecting the model's ability to collaboratively learn and predict highly correlated labels. However, previous multi-label batch selection methods have not accounted for these dynamic, uncertainty-based label correlations \cite{mllbalance,mllour}.

To tackle the two issues, we proposed an uncertainty-based multi-label batch selection method, which considers both current confidence and fine-grained variance in sliding windows, and leverages dynamic label correlations to emphasize the importance of uncertain samples during the training.
Specifically, we propose a new absolute difference-based measure to average the changes between adjacent predictions within the sliding window for each label, which reflects the reliabilities of the current prediction and fine-grained fluctuations between successive predictions.
Based on individual label uncertainty, we derive the dynamic uncertainty-based label correlations in each epoch, and prioritize samples whose uncertainty is synergistically expressed in more labels for mini-batch selection.

The main contributions of this paper are as follows:
\begin{itemize}
    \item \textbf{Label Uncertainty Estimation:} Our method provides a comprehensive assessment that integrates both present uncertainty and fine-grained changes in recent predictions to evaluate uncertainty for each label.
    \item \textbf{Sample Uncertainty Estimation:} Our method leverages dynamic uncertainty-based label correlations to guide the sample uncertainty assessment, emphasizing instances with higher uncertainty synergistically expressed in more labels in each epoch.
   
   \item \textbf{Effectiveness and Universality:} Our method achieves the most performance improvement compared with five competitors. In addition, the superiority of our method remains consistent across various deep multi-label learning models and datasets from different domains.
\end{itemize}

\section{Related Work \label{related work}}

\subsection{Multi-Label Classification}

Initially, multi-label classifiers adapt conventional machine learning techniques, such as neighborhood-based classifier \cite{MLKNN}, decision tree \cite{ML-DT}, and kernel method \cite{ML-SVM1}, to handle multi-label data. Alternatively, another solution converts multi-label classification into multiple single-label problems, which are solved by well-studied single-label models \cite{mllreview}. Representative strategies include individual label \cite{BR}, label pair \cite{COCOA}, label subset \cite{RAkEL}, and label chain \cite{ECCRU3}-based transformations.


Recently, deep neural networks (DNNs) have emerged as a highly successful technique for solving multi-label classification tasks.
Deep embedding-based methods effectively align feature and label spaces using DNNs. 
For example, \textit{C2AE} \cite{C2AE} embeds features and labels into a deep latent space with a label-correlation sensitive loss function. \textit{MPVAE} \cite{MPVAE} aligns probabilistic embeddings of labels and features, using a decoder to model their joint distribution. 
Some DNNs focus on capturing label correlations or learning label-specific features. 
For example, \textit{PACA} \cite{PACA} learns label prototypes and metrics in a latent space regulated by label correlations, while \textit{HOT-VAE} \cite{HOTVAE} uses attention to capture high-order label correlations adaptively. \textit{CLIF} \cite{CLIF} integrates label semantics with label-specific feature extraction using a graph autoencoder, and \textit{DELA} \cite{DELA} employs perturbation-based techniques for stable label-specific features within a probabilistic framework.
All deep multi-label classification models utilize randomly selected mini-batches to optimize the model, which fails to emphasize the crucial instances during the learning procedure.

\begin{table*}[!h]
\caption{The summary of uncertainty-based batch selection methods for single label data and multi-label batch selection approaches}
\label{table:batchselectmethod}
\centering
\resizebox{1.0\textwidth}{!}{
\begin{tabular}{ccccc}
\toprule
 Method & Criteria & Uncertainty Measure & Label correlation & Datatype \\ \midrule
\textit{Active Bias} \cite{activebias} & uncertainty & variance of entire prediction history  & - &single-label \\ 
\textit{Recent Bias} \cite{RecencyBias}  & uncertainty & entropy of recent predictions & -& single-label \\ 
\textit{Balance} \cite{mllbalance} & imbalance  &  \xmark  & \xmark & multi-label \\ 
\textit{Hard Imbalance} \cite{mllour} &  hardness \& imbalance  & \xmark & \xmark & multi-label \\ 
\textbf{\textit{Ours}} \ & uncertainty & \begin{tabular}[c]{@{}c@{}} fine-grained fluctuations of recent \\ windows and current prediction\end{tabular} & \cmark & multi-label \\  \bottomrule 
\end{tabular}
}
\label{sota_batch_method}
\end{table*}

\subsection{Batch Selection}
Recent research emphasizes that the performance of DNNs depends on the selection of mini-batch samples \cite{OHEM,batch_quality2,Ada-boundary,activebias}. 
Batch selection has been used in various learning strategies such as reinforcement learning \cite{reinforcementlearning}, curriculum learning \cite{Curriculumlearning}, and active learning \cite{mllactive1,mllactive2}, as well as in different learning tasks like classification \cite{Onlinebatch,Ada-boundary} and sample labeling \cite{mllactive1}.

Sample difficulty plays a crucial role in mini-batch selection. Two opposing strategies—preferring easy or hard samples—are effective in different scenarios. Prioritizing easy samples helps resist outliers and noisy labels but may slow training due to smaller gradients \cite{easy,Selfie}. In contrast, focusing on hard samples accelerates training but can cause overfitting and poor generalization \cite{Onlinebatch}. Additionally, some heuristic batch selection methods have proven effective in single-label datasets. 
\textit{Ada-Boundary} \cite{Ada-boundary} focuses on moderately challenging samples near the decision boundary to optimize learning progress. 
\textit{Active Bias} \cite{activebias} uses uncertainty-based sampling, prioritizing uncertain samples for the next batch. It maintains a history queue storing all previous predictions and measures uncertainty by computing the prediction variance. 
\textit{Recency Bias} \cite{RecencyBias} also measures the variance of recent predictions within a fixed-sized sliding window, eliminating the impact of outdated predictions on the uncertainty estimation. For multi-label data, \textit{Balance} \cite{mllbalance} adjusts batches to match desired label distributions, balancing over- and under-represented labels by sampling and weighting instances. \textit{Hard Imbalance} \cite{mllour} prioritizes samples with high losses and imbalanced labels based on cross-entropy loss and label imbalance. In Table \ref{sota_batch_method}, we summarize several SOTA batch selection methods in single-label or multi-label data.


\section{Proposed Method}
In this section, we first compute the uncertainty for each label from both the current epoch and the recent historical window perspectives. Next, we derive the sample uncertainty by considering the dynamic uncertainty-based label correlation. Finally, we calculate the probability of selecting the next batch of samples based on their uncertainty-based weights.

\subsection{Problem Formulation}

Let $\mathcal{D} = \left \{(\mathbf{x}_i, \mathbf{y}_i) {|}_{i=1}^{n} \right \}$ be a multi-label dataset containing $n$ instances, where \( \mathbf{x}_i \in \mathbb{R}^d \) and  \( \mathbf{y}_i = [y_{i1}, y_{i2}, \ldots, y_{iq}] \in \{0, 1\}^q \) are the feature and label vectors of $i$-th instance, respectively. Let $\mathcal{Y} = \{l_1, l_2, \ldots, l_q\}$ be the label set, $y_{ij} = 1$ indicates $i$-th instance relevant to $l_j$ and \( y_{ij} = 0 \) otherwise.
Formally, multi-label classifiers aim to learn from dataset $\mathcal{D}$ a function $f(\cdot): \mathbb{R}^d \rightarrow  \{0, 1\}^q $ that maps the input features to output labels.
For training deep multi-label learning models, selecting a mini-batch $\mathcal{B}=\left \{(\mathbf{x}_i, \mathbf{y}_i) {|}_{i=1}^{b}\right \} \in \mathcal{D}$ is necessary to update their parameters (weights) due to efficiency and machine memory constraints.


\subsection{Label Uncertainty \label{labeluncertain}} 

For a sample $\mathbf{x}_{i}$, the probability of label $l_j$ given by the model at epoch $t$ is defined as $\hat{y}^t_{ij} = P(y_{ij}=1|\mathbf{x}_{i},\theta_t)$, where $\hat{y}^t_{ij} \in [0,1]$ with larger values indicating $\mathbf{x}_{i}$ is more likely relevant to label $l_j$, $\theta_t$ are the parameters of the deep learning model at epoch $t$.
We use entropy to measure the uncertainty of each label prediction at the current epoch:
\begin{equation}
e^t_{ij} = -\left( \hat{y}^t_{ij} \log_2\hat{y}^t_{ij} + (1-\hat{y}^t_{ij}) \log_2(1-\hat{y}^t_{ij}) \right)
\label{empiricaluncertain}
\end{equation}
The value of \( e^t_{ij} \) is in the range of [0,1], with the maximum value obtained when \(\hat{y}_{ij} = 1/2\). 
A larger \(e^t_{ij}\) indicates lower confidence (higher uncertainty) of the prediction for $l_j$ label at epoch $t$.

\begin{figure}[!h]
    \centering
    \includegraphics[width=1.0\linewidth]{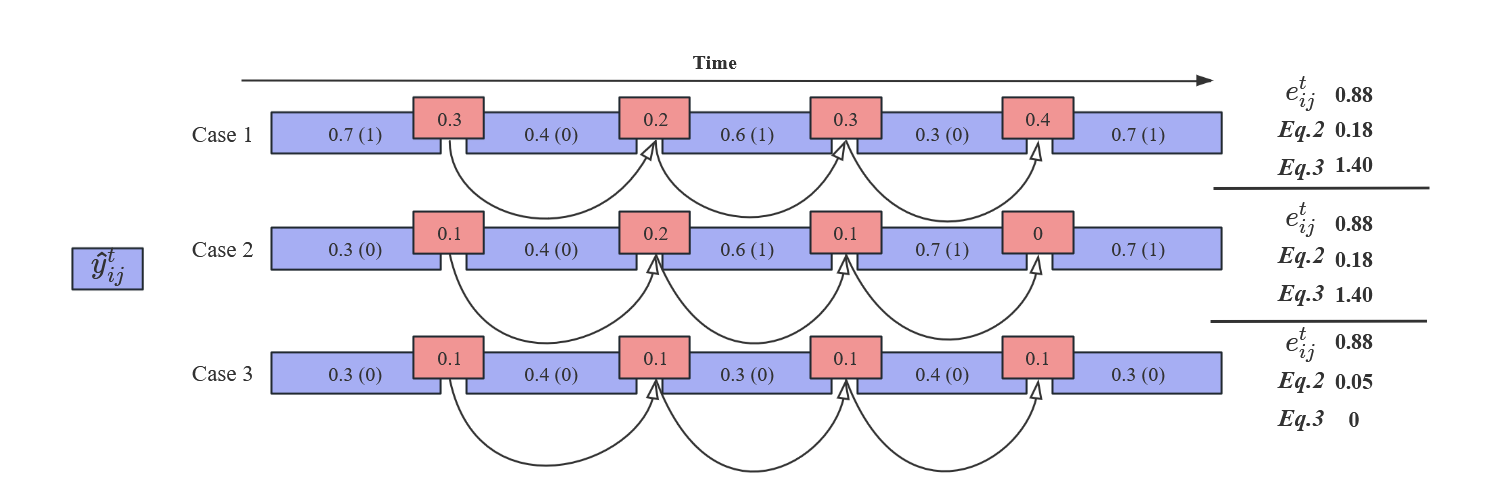}
    \caption{The different uncertainty measurement methods across three cases for historical predictions of the label.}
    \label{fig:twoscenarios}
\end{figure}

Merely considering the uncertainty at the current epoch is not sufficient. As shown in Figure \ref{fig:twoscenarios}, the uncertainty obtained by Eq. \eqref{empiricaluncertain} is the same in three different cases. However, by tracing through several historical windows, we observe that three cases exhibit different historical predictive trends. 
Inspired by \cite{RecencyBias}, we consider the uncertainty based on historical window prediction to explain the finer details that Eq .\eqref{empiricaluncertain} cannot distinguish.
Let \(H_{ij}^t=\{\hat{y}_{ij}^{t-T+1}, \hat{y}_{ij}^{t-T+2}, \ldots, \hat{y}_{ij}^t\}\) be a prediction history queue corresponding to a sliding window of size \(T\) at epoch $t$.
There are two traditional measures to evaluate the uncertainty of $H_{ij}^t$, namely prediction variance \cite{activebias}:
\begin{equation}
std(H_{ij}^t) = \displaystyle \sqrt{var(H_{ij}^t)+\frac{var(H_{ij}^t)^2}{\left | H^t_{ij}\right|-1 }}
\label{uncertain1}
\end{equation}
where $var(H_{ij}^t)$ is the prediction variance estimated by history $H_{ij}^t$, 
and entropy-based uncertainty \cite{RecencyBias}:
\begin{align}
entropy(H_{ij}^t) & = - \sum_{c \in \{0,1\}}  P(y_{ij} = c| \mathbf{x}_i) \log_2 P(y_{ij} = c| \mathbf{x}_i), \nonumber\\
& P(y_{ij} = c| \mathbf{x}_i)  = \frac{\sum_{\hat{y} \in H_{ij}^t}  \llbracket \widetilde{y}=c \rrbracket}{T}
\label{uncertain2}
\end{align}
where $\widetilde{y} \in \{0,1\}$ is the binary prediction based on its predicting probability $\hat{y}$, $\llbracket \cdot \rrbracket$ is an indicator function that returns 1 if the input is true and 0 otherwise. 
As shown in Figure \ref{fig:twoscenarios}, when considering the recent five predictions (i.e., $T$=5), 
for the first two cases, if we use Eq. \eqref{uncertain1} or Eq. \eqref{uncertain2} to measure uncertainty, the uncertainty in these two cases will be the same. However, in case 1, the model's predictions exhibit greater volatility (indicating higher uncertainty), 
whereas case 2 can be seen as a model prediction with a certain trend. 
Therefore, to capture prediction fluctuations within the window at a finer granularity, we decided to extract information from the differences in queue $H_{ij}^t$. In detail, we define \( d_{ij}^t \) as the mean of the absolute differences between adjacent predictions within $H_{ij}^t$, denoting the uncertainty of the historical period:
\begin{equation}
d_{ij}^t = \frac{1}{T-1} \sum_{h=1}^{T-1} \left| \hat{y}_{ij}^{t-h+1} - \hat{y}_{ij}^{t-h} \right|
\label{historicaluncertain}
\end{equation}
where the values of \( d_{ij}^t \) range between 0 and 1. The larger the value of $d_{ij}^t$, the greater the uncertainty in the historical period. 

Based on Eq. \eqref{empiricaluncertain} and Eq. \eqref{historicaluncertain}, we define the uncertainty \(u^t_{ij}\) of \(x_i\) regarding \(l_j\) at epoch $t$ that combines uncertainties of the current prediction and recent variant trend:
\begin{equation}
u^t_{ij} = \lambda_1 d^t_{ij} + (1-\lambda_1) e^t_{ij}
\label{uncertain}
\end{equation}
where \(\lambda_1\) is trade-off parameters determining the importance of two factors.
$u^t_{ij}$ of all training instances and labels compose the uncertainty matrix $\mathbf{U}^t \in \mathbb{R}^{n \times q}$ \footnote{In the following text, $\mathbf{U}$ without the superscript \( t \) indicates the current value to avoid reading difficulties caused by excessive symbols.}. 

\subsection{Label Correlation Guided Sample Uncertainty}
Considering each sample, assuming each label is completely independent, we can directly sum the uncertainties for every label, i.e., $\sum_{j=1}^q u^t_{ij}$, as a measure of the sample's uncertainty. 
However, a key characteristic of multi-label data is the existence of label correlations. 
During model training, the uncertainty of each label varies and changes dynamically. Ideally, this uncertainty should gradually approach zero. 
Furthermore, we hypothesize that there is a correlation between the uncertainties of different labels. This correlation may arise due to shared underlying factors that affect multiple labels simultaneously. For instance, in a multi-label classification problem, certain features might influence several labels, leading to simultaneous high uncertainty when these features are ambiguous or conflicting. 
This can occur when different labels share common subspaces or dependencies, where uncertainty in one label can imply uncertainty in others. Understanding and quantifying this correlation can provide valuable insights into the overall uncertainty of samples. For example, if we observe that high uncertainty in one label often coincides with high uncertainty in others, we can infer that these samples are inherently more challenging and may require more attention during training or evaluation. 

First, a discrete distribution is formed using each column of \(\mathbf{U}\) (denoted as $\mathbf{u}_{\cdot j}$) by placing the $u_{ij}$ in \(\tau\) bins of width \( 1/\tau\). For two labels \( l_a \) and \( l_b \), mutual information \( C_{ab} \) is defined as:
\begin{equation}
C_{ab} =\displaystyle\sum_{u^\tau_{\cdot a} \in \mathbf{u}_{\cdot a}} \displaystyle\sum_{u^\tau_{\cdot b} \in \mathbf{u}_{\cdot b}}p(u^\tau_{\cdot a},u^\tau_{\cdot b}) \log \left( \frac{p(u^\tau_{\cdot a}, u^\tau_{\cdot b})}{p(u^\tau_{\cdot a}) p(u^\tau_{\cdot b})} \right) 
\label{mutualinformation}
\end{equation}
where \(p(u^\tau_{\cdot a},u^\tau_{\cdot b})\) represents the joint probability distribution of \( u^\tau_{\cdot a} \) and \( u^\tau_{\cdot b} \), where \( u^\tau_{\cdot a} \) denotes the bin in which the value \( u_{ij} \) from the $\mathbf{u}_{\cdot j}$, with \(\tau\) bins in total. Similarly, \( p(u^\tau_{\cdot a}) \) and \( p(u^\tau_{\cdot b}) \) are the marginal probability distributions of the values in these bins for the labels \( l_a \) and \( l_b \), respectively \footnote{The detailed calculation process can be found in Appendix A}. 
The larger the mutual information \( C_{ab} \), the stronger the dependency between the two labels, indicating more shared information between them. Based on Eq. \eqref{mutualinformation}, we obtain the positive definite symmetric matrix $\mathbf{C}$, where the diagonal elements are defined as 1. 
By combining label correlation with uncertainty, we re-obtain an uncertainty matrix $\bar{\mathbf{U}}$:
\begin{equation}
\bar{\mathbf{U}} = \mathbf{U} \cdot \mathbf{C}
\label{reuncertaintymatrix}
\end{equation}
Finally, we define the uncertainty weight vector \(\mathbf{w}=[w_1, w_2, \ldots, w_n] \in \mathbb{R}^n\). For the $i$-th sample, the uncertainty weight \(w_i\) is defined as:
\begin{equation}
w_i = \sum_{j=1}^{q}u_{ij}
\label{uncertainweight}
\end{equation}
For all samples, the $\mathbf{w}$ is normalized to the range [0, 1].

\begin{figure}[h]
    \centering
    \includegraphics[width=1.0\linewidth]{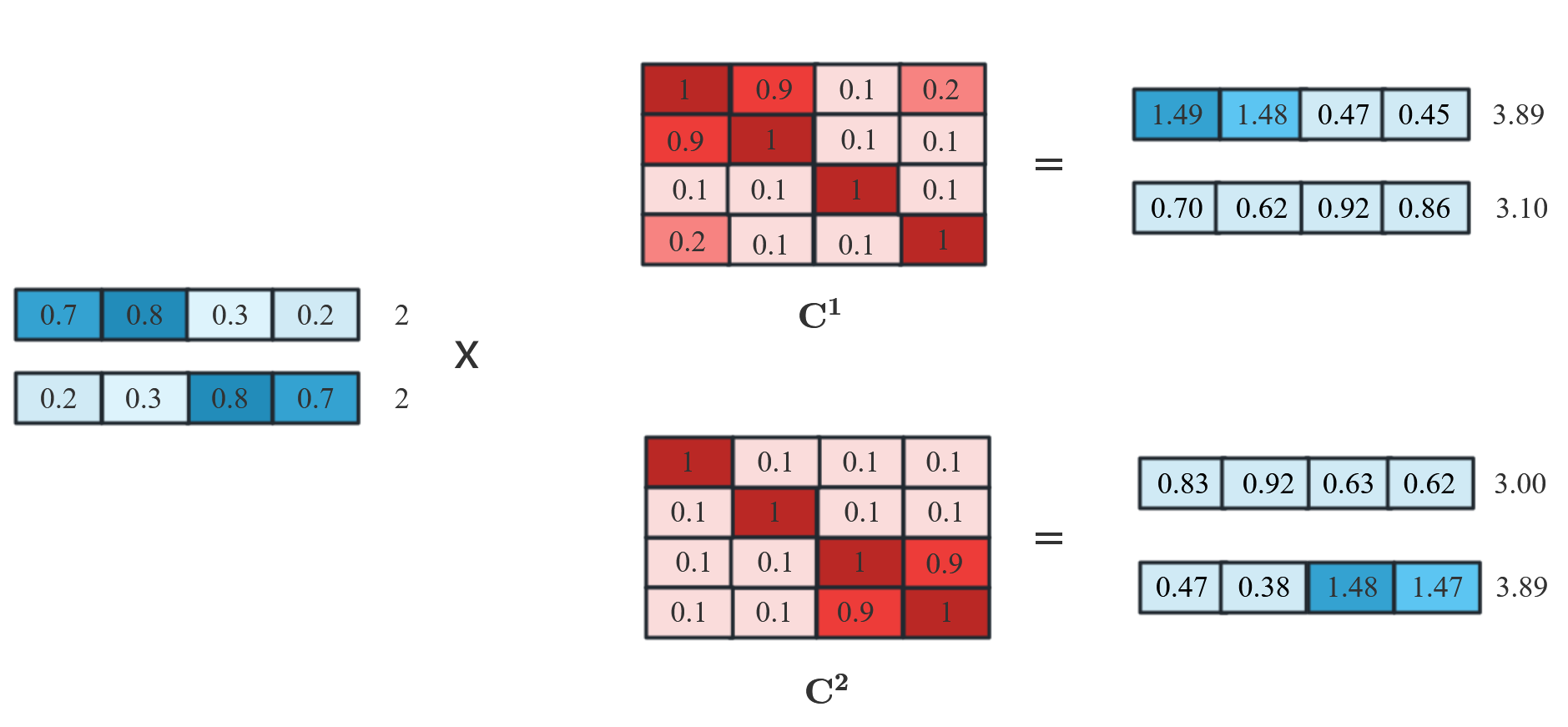}
    \caption{If we directly add up the uncertainty of each label, the total uncertainty of the first sample is equal to that of the second sample. However, after introducing the mutual information matrix \(\mathbf{C}\), the recalculated uncertainty matrix shows that under the influence of the first mutual information matrix, the total uncertainty of the first sample (3.89) is greater than the uncertainty of the second sample (3.10), whereas under the influence of the second mutual information matrix, the total uncertainty of the first sample (3.00) is less than the uncertainty of the second sample (3.89).}
    \label{fig:uncertainty_corelation}
\end{figure}

\subsection{Selection probability}
Motivated by \cite{Ada-boundary,RecencyBias}, we exponentially decay the sampling probability of the $i$-th sample based on its uncertainty weight \(w_i\).
In detail, we utilize a quantization method to reduce sampling probabilities, with the quantization index derived from a simple quantizer \( Q(z) \) as follows:
\begin{equation}
Q(z) = \lfloor (1-z) /\Delta\rfloor
\label{quantization}
\end{equation}
where \(\Delta\) is the quantization step size, defined as \(1/n\), with \(n\) representing the total number of samples. This ensures that the quantization index is bounded by \(n\). A crucial component of our approach is the selection pressure \(s_t\), which controls the distribution of sampling probabilities over time. The sampling probability \(P(x_i|\mathcal{D}, w_i, n, s_t)\) is then defined as:
\begin{equation}
P(x_i|\mathcal{D}, w_i, n, s_t) = \frac{1/\exp(\log(s_t)/n)^{Q(w_i)}}{\sum_{i=1}^{n} 1/\exp(\log(s_t)/n)^{Q(w_i)}}
\label{probability}
\end{equation}
The higher the uncertainty, the smaller the quantization index. Therefore, a higher selection probability is assigned for more uncertain samples by Eq. \eqref{probability}. 
For $s_t$, to mitigate overfitting caused by using only a portion of the training data, we gradually increase the number of training samples as training progresses. This is achieved by exponentially decaying the selection pressure \(s_t\) using
\begin{equation}
s_t = s_{0} \left( \exp \left( \log(1/s_{0})/(t_{end} - t_{start}) \right) \right)^{t_{now}-t_{start}}
\end{equation}
At each epoch \(t_{now}\) from \(t_{start}\) to \(t_{end}\), the selection pressure \(s_t\) exponentially decreases from \(s_0\) to 1. Because this technique gradually reduces the sampling probability gap between the most and the least uncertain samples, more diverse samples are selected for the next mini-batch at a later epoch. When the selection pressure \(s_t\) becomes 1, the most and the least uncertain samples are sampled more uniformly.
The convergence guarantee of Algorithm \ref{al:uncertainbatch} is discussed in Appendix B.

\begin{algorithm}[!h]
\KwIn{training set $\mathcal{D}$ , \textit{epochs}, batch size: $b$, initial select presure: $s_{0}$, warm period:$\gamma$, Model: $\Theta$ }
Initialize the $\mathbf{U}$,$\mathbf{C}$,$P$\;
\For{$t=1$ to epochs}{  
    \If{$t>\gamma$}
    {
    $s_t$ $\gets$ Decay Pressure($s_{0}$,$t$)\;
    
    Update $\mathbf{C}$ by Eq. \eqref{mutualinformation}\;

    \For{$i=1$ to $n$}{  
    $w_i$ $\gets$ Compute uncertain weight\;
    $P$($x_i$) $\gets$  Compute Prob($s_t$,$w_i$)
    }
    }
    \textbf{Model training}\;
    \For{$i=1 \gets$ n/b}{
    \eIf{$t<\gamma$}
    {
    $\mathcal{B}$=$\left \{(x_i,Y_i){|}_{i=1}^{b} \right \}$ $\leftarrow$ Random selection
    }
    {
    $\mathcal{B}$=$\left \{(x_i,Y_i){|}_{i=1}^{b} \right \}$ $\leftarrow$ $P$($x_i$)
    }
     Forward\; 
     Update $\mathbf{U}$ by Eq. \eqref{empiricaluncertain} \eqref{historicaluncertain} \eqref{uncertain} \eqref{mutualinformation} \eqref{reuncertaintymatrix}\;
     Calculate loss and Backward\; 
     Optimize $\Theta$
}
}
\caption{Training by Uncertain Batch Selection}
\label{al:uncertainbatch}
\end{algorithm}


\section{Experiments and Analysis}

\subsection{Experiment Setup}

\subsubsection{Datasets}

Characteristics and imbalance levels of these datasets are detailed in Table \ref{ta:mld}, including \textit{Card}, the mean labels per instance associated, and \textit{Dens}, the ratio of \textit{Card} to the overall label count.

\begin{table}[h]
\centering
\caption{Datasets}
\label{ta:mld}
\resizebox{0.45\textwidth}{!}{
\begin{tabular}{cccccccccc}
\toprule
name & $n$ & $d$ & $q$ & \textit{Card} & \textit{Dens}  & domain \\  
\midrule
scene & 2407 & 294 & 6 & 1.07 & 0.18 & images \\
yeast & 2417 & 103 & 14 & 4.24 & 0.30 & biology \\
Corel5k & 5000 & 499 & 374 & 3.52 & 0.01 & images \\
rcv1subset1 & 6000 & 944 & 101 & 2.88 & 0.03 & text \\
rcv1subset2 & 6000 & 944 & 101 & 2.63 & 0.03 & text \\
rcv1subset3 & 6000 & 944 & 101 & 2.61 & 0.03 & text \\
yahoo-Arts & 7484 & 2314 & 25 & 1.67 & 0.07 & text \\
yahoo-Business & 11214 & 2192 & 28 & 1.47 & 0.06 & text \\
bibtex & 7395 & 1836 & 159 & 2.40 & 0.02 & text \\
tmc2007 & 28596 & 490 & 22 & 2.15 & 0.10 & text \\
enron & 1702 & 1001 & 53 & 3.38 & 0.06 & text \\
cal500 & 502 & 68 & 174 & 26.04 & 0.15 & music \\
LLOG-F & 1460 & 1004 & 75 & 15.93 & 0.21 & text \\
\bottomrule
\end{tabular}
}
\end{table}


\subsubsection{Comparison method}
\begin{table*}[!h]
\centering
\caption{The Macro-AUC results of batch selection methods under different models, where the best results are highlighted by \textbf{boldface}.}  
\label{ta:Macro-AUC}
\resizebox{\textwidth}{!}{
\begin{tabular}{ccccccc|cccccc|cccccc}
\toprule
\multirow{2}{*}{Dataset} & \multicolumn{6}{c}{\textit{MPVAE}} & \multicolumn{6}{c}{\textit{CLIF}} & \multicolumn{6}{c}{\textit{DELA}} \\ \cmidrule(lr){2-7} \cmidrule(lr){8-13} \cmidrule(lr){14-19}
& \textit{Random} & \textit{Balance} & \textit{Active} & \textit{Recent} & \textit{Hard} & \textit{Ours} & \textit{Random} & \textit{Balance} & \textit{Active} & \textit{Recent} & \textit{Hard} & \textit{Ours} & \textit{Random} & \textit{Balance} & \textit{Active} & \textit{Recent} & \textit{Hard} & \textit{Ours} \\ \midrule
scene & 0.9303(6) & 0.9355(4) & 0.9349(5) & 0.9368(2) & 0.9361(3) & \textbf{0.9396(1)}&
0.9418(6) & 0.9442(4) & 0.9437(5) & 0.9446(3) & 0.9454(2) & \textbf{0.9476(1)} 
& 0.9405(6) & 0.9432(5) & 0.9458(3) & \textbf{0.9476(1)} & 0.9449(4) & 0.9465(2) \\

yeast & 0.7065(5) & 0.7068(4) & 0.7057(6) & 0.7082(3) & 0.7100(2) & \textbf{0.7126(1)}
& 0.7107(5) & 0.7077(6) & 0.7164(4) & 0.7195(2) & 0.7191(3) & \textbf{0.7222(1)} 
& 0.7006(4) & 0.6972(6) & 0.6996(5) & 0.7036(3) & 0.7121(2) & \textbf{0.7132(1)} \\

Corel5k & 0.6365(2) & 0.6243(6) & 0.6303(5) & 0.6342(4) & 0.6361(3) & \textbf{0.6378(1)}
& 0.7664(3) & 0.7625(6) & 0.7650(4) & 0.7648(5) & 0.7683(2) & \textbf{0.7689(1)} 
& 0.7626(2) & 0.7583(6) & 0.7594(5) & 0.7618(4) & 0.7621(3) & \textbf{0.7664(1)} \\

rcv1subset1 & 0.8496(6) & 0.8526(5) & 0.8543(4) & 0.8566(3) & 0.8589(2) & \textbf{0.8596(1)}
& 0.9221(6) & 0.9262(5) & 0.9281(4) & 0.9296(3) & 0.9307(2) & \textbf{0.9324(1)} 
& 0.9179(5) & 0.9169(6) & 0.9187(4) & 0.9190(3) & 0.9194(2) & \textbf{0.9204(1)} \\

rcv1subset2 & 0.8257(5) & 0.8242(6) & 0.8275(4) & 0.8296(2) & 0.8277(3) & \textbf{0.8302(1)} 
& 0.9279(6) & 0.9316(5) & 0.9326(4) & 0.9338(2) & 0.9329(3) & \textbf{0.9345(1)} 
& 0.9203(6) & 0.9220(2) & 0.9213(5) & 0.9216(4) & 0.9220(2) & \textbf{0.9226(1)} \\

rcv1subset3 & 0.8136(6) & 0.8271(4) & 0.8236(5) & 0.8276(3) & 0.8298(2) & \textbf{0.8316(1)}
& 0.9268(6) & 0.9277(5) & 0.9282(3) & \textbf{0.9308(1)} & 0.9279(4) & 0.9306(2) 
& 0.9175(6) & 0.9178(5) & 0.9185(3) & 0.9189(2) & 0.9183(4) & \textbf{0.9192(1)} \\

yahoo-Business1 & 0.7478(3) & 0.7493(2) & 0.7446(6) & 0.7472(4) & 0.7455(5) & \textbf{0.7497(1)}
& 0.7801(6) & 0.7926(3) & 0.7862(5) & 0.7884(4) & 0.7940(2) & \textbf{0.8077(1)} 
& 0.7979(5) & 0.7972(6) & 0.8026(3) & 0.8039(2) & 0.8010(4) & \textbf{0.8086(1)} \\

yahoo-Arts1 & 0.7208(4) & 0.7126(6) & 0.7225(3) & 0.7269(2) & 0.7187(5) & \textbf{0.7302(1)}
& 0.7580(4) & 0.7591(3) & 0.7598(2) & \textbf{0.7599(1)} & 0.7535(6) & 0.7562(5) 
& 0.7407(5) & 0.7395(6) & 0.7422(4) & 0.7460(2) & 0.7444(3) & \textbf{0.7480(1)} \\

bibtex & 0.8594(3) & 0.8454(5) & 0.8432(6) & 0.8497(4) & \textbf{0.8674(1)} & 0.8662(2)
& 0.9013(2) & 0.8975(6) & 0.8996(4) & 0.8982(5) & 0.9011(3) & \textbf{0.9059(1)} 
& \textbf{0.9062(1)} & 0.8974(6) & 0.9046(5) & 0.9052(4) & 0.9057(3) & 0.9060(2) \\

tmc2007 & 0.8697(5) & 0.8669(6) & 0.8698(4) & 0.8722(2) & 0.8711(3) & \textbf{0.8732(1)} 
& 0.9048(5) & 0.9053(3) & 0.9051(4) & 0.9046(6) & 0.9059(2) & \textbf{0.9062(1)} 
& 0.9121(4) & 0.9087(6) & 0.9145(3) & 0.9162(2) & 0.9120(5) & \textbf{0.9179(1)} \\

enron & 0.6690(6) & \textbf{0.6799(1)} & 0.6693(5) & 0.6735(3) & 0.6698(4) & 0.6764(2)
& 0.7700(6) & 0.7744(5) & 0.7753(4) & \textbf{0.7782(1)} & 0.7763(3) & 0.7764(2) 
& 0.7727(6) & 0.7732(5) & 0.7748(3) & \textbf{0.7806(1)} & 0.7748(3) & 0.7754(2) \\

cal500 & 0.5881(3) & 0.5868(4) & 0.5806(6) & 0.5814(5) & 0.5972(2) & \textbf{0.5985(1)}
& 0.5901(4) & 0.5949(2) & 0.5843(6) & 0.5885(5) & 0.5932(3) & \textbf{0.6054(1)} 
& 0.5933(2) & 0.5761(6) & 0.5872(5) & 0.5914(4) & 0.5927(3) & \textbf{0.5954(1)} \\

LLOG-F & 0.6375(6) & \textbf{0.6560(1)} & 0.6432(4) & 0.6447(3) & 0.6419(5) & 0.6452(2) 
& 0.7659(6) & 0.7686(4) & 0.7682(5) & 0.7716(2) & 0.7703(3) & \textbf{0.7721(1)} 
& 0.7909(6) & 0.7911(5) & 0.7916(4) & \textbf{0.7935(1)} & 0.7930(2) & 0.7924(3) \\ \midrule

Avg (Rank)& 4.62 & 4.15 & 4.85 & 3.08 & 3.08 & \textbf{1.23} & 5.00 & 4.38 & 4.15 & 3.08 & 2.92 & \textbf{1.46} & 4.46 & 5.38 & 4.00 & 2.54 & 3.08 & \textbf{1.38} \\\bottomrule

\end{tabular}}
\end{table*}
We compare the proposed method with the following batch selection methods: \textit{Random}, \textit{Balance} \cite{mllbalance}, \textit{Active} \cite{activebias}, \textit{Recent} \cite{RecencyBias} and \textit{Hard} \cite{mllour}. 
Among them, \textit{Active} and \textit{Recent} are originally designed for single-label scenarios. To adapt them for multi-label data, we calculate the uncertainty of a sample by summing the uncertainty of each label. Details of the batch selection methods and parameter settings can be found in the related work and Appendix Section C.

\subsubsection{Evaluation Metrics}
To evaluate the effectiveness of the batch selection method in multi-label classification, we use three common metrics: Macro-AUC, Ranking Loss, and Hamming Loss. Please refer to \cite{mllreview} for detailed definitions of these metrics.

\subsubsection{Base Classifier and Implementation Details}
We use three multi-label deep models as base classifiers, namely \textit{MPVAE} \cite{MPVAE}, \textit{CLIF} \cite{CLIF}, and \textit{DELA} \cite{DELA}. 
We configure each model precisely according to the parameter specifications, encompassing layer sizes, activation functions, and other intricate details, outlined in the corresponding original research papers and source codes. In terms of optimization, we utilize the Adam optimizer with a batch size of 128, a weight decay of 1e-4, and momentum values of 0.999 and 0.9. 
For the hyperparameter settings, we fix \(\lambda_1\) at 0.5, set the selection pressure \(s_t\) initially to 100, and use a sliding window size \(T\) of 5. 
During the first 5 epochs, we employ a warm-up phase to initialize the historical predictions for each instance's label.
For experiments, we adopt stratified five-fold cross-validation \cite{stratification} to evaluate the aforementioned models. 
In each fold, we document the test set results achieved at the epoch that yields the best performance on the validation set. 
All experiments in this work are conducted on a machine with NVIDIA A5000 GPU and Intel Xeon i9-10900 processor.
Our code is publicly available on GitHub repository 
\url{https://github.com/CquptZA/Uncertainty_Batch}.

\subsection{Experimental Results and Analysis}
\subsubsection{Results}
Table \ref{ta:Macro-AUC} presents the average Macro-AUC comparison of six different batch selection methods within \textit{MPVAE}, \textit{CLIF}, and \textit{DELA}. 
The results of other metrics and the Wilcoxon signed-ranks test are shown in Appendix Section D.
Our batch selection methods consistently achieve the best performance in most datasets. This advantage is particularly evident in Corel5k, rcv1subset1, cal500, and bibtex datasets with larger scales, high label dimensions, or suffering significant imbalance issues. Additionally, our batch selection method performs exceptionally well across different models (\textit{MPVAE}, \textit{CLIF}, \textit{DELA}), demonstrating their versatility and robustness. 
\textit{Hard} is usually the runner-up, indicating that selecting samples based on their relevance to the learning objectives and prioritizing more informative and challenging samples is beneficial for the classifier. 
\textit{Active} batch selection, which considers uncertainty over the entire cumulative history and directly accumulates the uncertainty of all labels as the sample uncertainty, outperforms the baseline in most datasets. 
Similarly, \textit{Recent} batch selection considers uncertainty within the latest sliding windows and directly accumulates the uncertainty of all labels as the sample uncertainty. 
While it generally outperforms the baseline across most datasets, it underperforms on datasets with many labels. 
Although experiments on smaller datasets (such as yeast and enron) indicate that the \textit{Balance} method outperforms the baseline, the effectiveness of \textit{Balance} has not been sufficiently demonstrated on the majority of datasets.

\subsubsection{Analysis} 
\begin{figure*}[!h]
    \centering
    \includegraphics[width=0.8\linewidth]{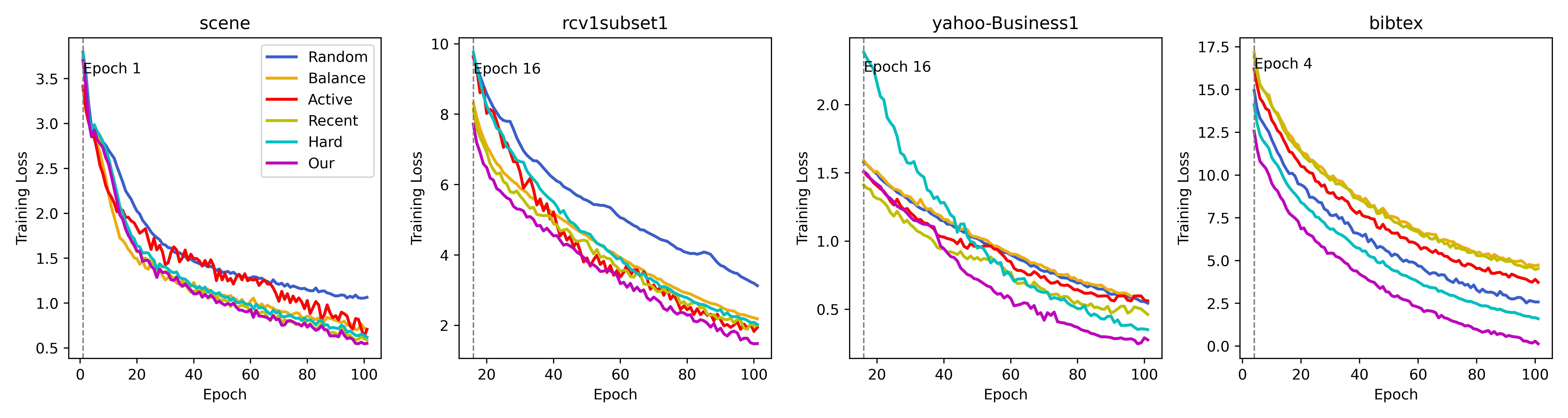}
    \caption{The convergence curves of five batch selection methods using \textit{CLIF}.}
    \label{fig:convergence}
\end{figure*} 
\begin{figure*}[!h]
    \centering
    \includegraphics[width=0.8\linewidth]{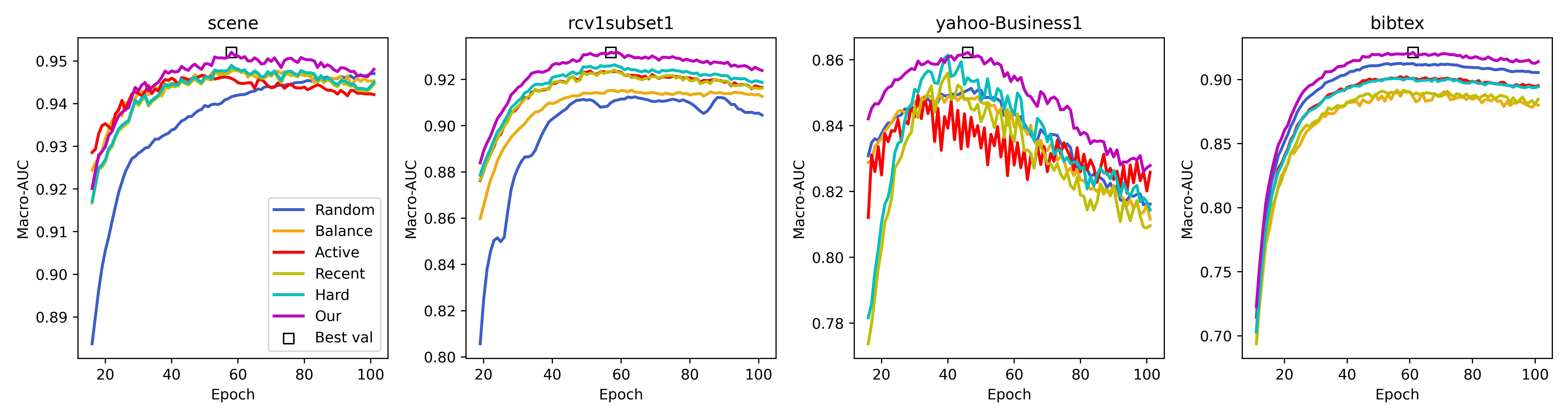}
    \caption{The Macro-AUC on validation set of five batch selection methods using \textit{CLIF}.}
    \label{fig:Macro-AUC}
\end{figure*} 
To conduct an in-depth analysis of the different batch selections, we plot the convergence curves for five different batching methods across four datasets in Figure \ref{fig:convergence}, and plot the Macro-AUC for each epoch on the validation set of the four datasets in Figure \ref{fig:Macro-AUC}.
\textit{Balance} determines batch assignments based on the label proportions in the original training set. This can lead to underrepresentation or overrepresentation of certain labels, which may cause overfitting in later stages of training, as observed in the bibtex and yahoo Business1 datasets. 
The \textit{Active} method focuses on moderately hard samples in the early stages of training, with a loss distribution between random and online batches. However, as the window continues to expand, predictions become outdated, and the proportion of low-loss, easy samples increases in the later stages, which may slow down the convergence.
Similarly, the \textit{Recent} method also emphasizes moderately hard samples, with a loss distribution similar to \textit{Active}. However, unlike \textit{Active}, the \textit{Recent} method uses a sliding window to dynamically update its selection of moderately hard samples throughout the training process. 
We find that traditional uncertainty-based batch selection methods exhibit less than ideal convergence speed and are prone to overfitting in datasets with large sample sizes (e.g., yahoo-Business) or high-dimensional label spaces (e.g., bibtex). This may be because, as sample size or label dimensions increase, identifying and prioritizing uncertain samples becomes more challenging, further complicating the batch selection process.
\textit{Hard} prioritizes high-loss samples associated with minority labels, resulting in more informative and challenging samples in each batch. This accelerates the learning process and leads to better generalization, but the \textit{Hard} method shows overfitting on certain datasets, such as yahoo-Business1. This could be due to the frequent selection of difficult samples in the later stages of training.
Due to its dynamic consideration of uncertainty using a sliding window, \textit{Our} batch selection typically results in better convergence by providing the model with moderately hard samples and avoiding the frequent selection of certain samples in the later stages through decaying selection pressure. 

More detailed empirical analyses, including computational complexity, ablation studies, and parameter sensitivity, are detained in Appendices E, F and G, respectively.   


\subsubsection{Uncertainty based Label Correlation Drift} 

\begin{figure}[h]
\centering
\subfigure[30-th epoch]{\includegraphics[width=0.14\textwidth]{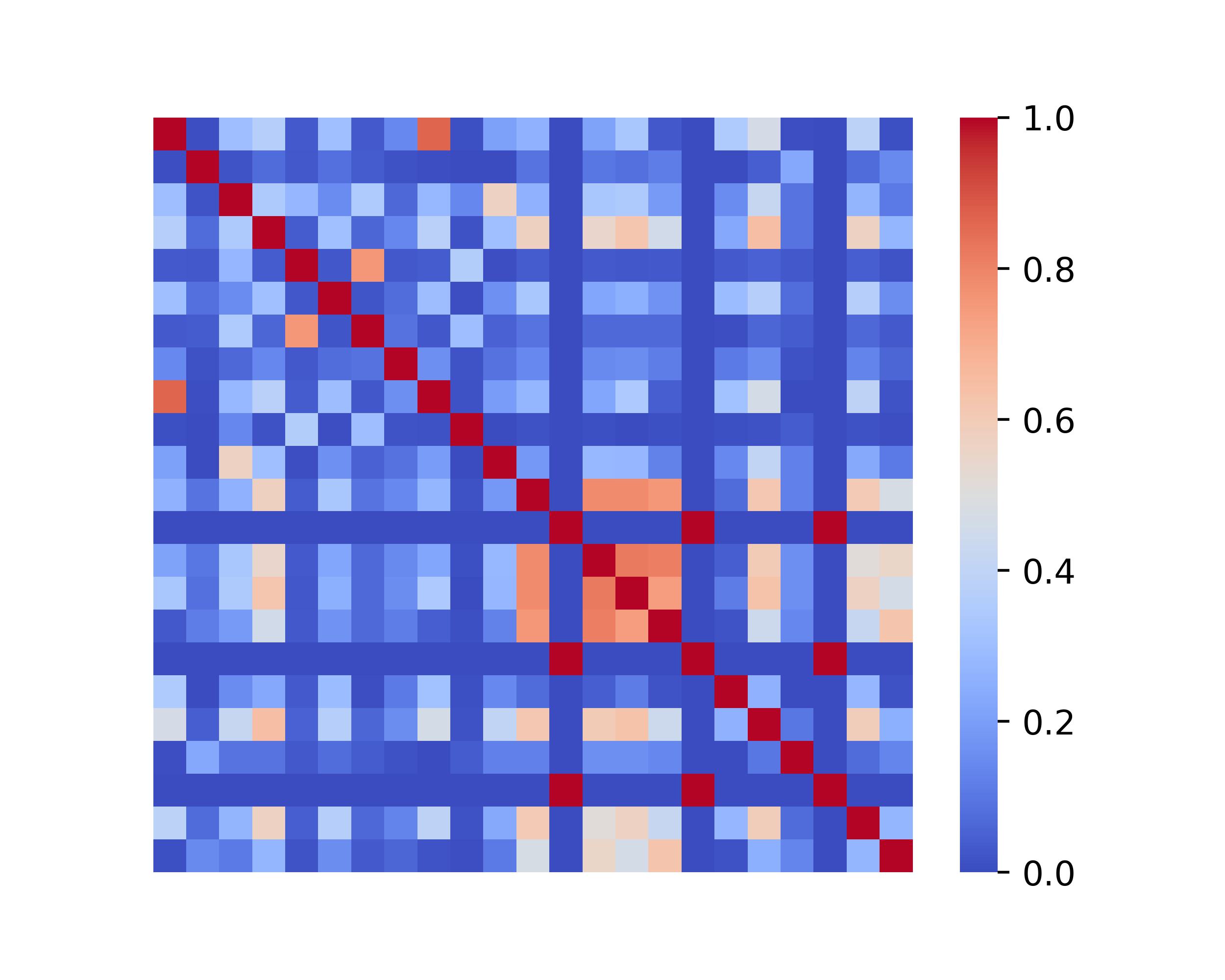}
\label{labelcor1}}
\subfigure[70-th epoch]{\includegraphics[width=0.14\textwidth]{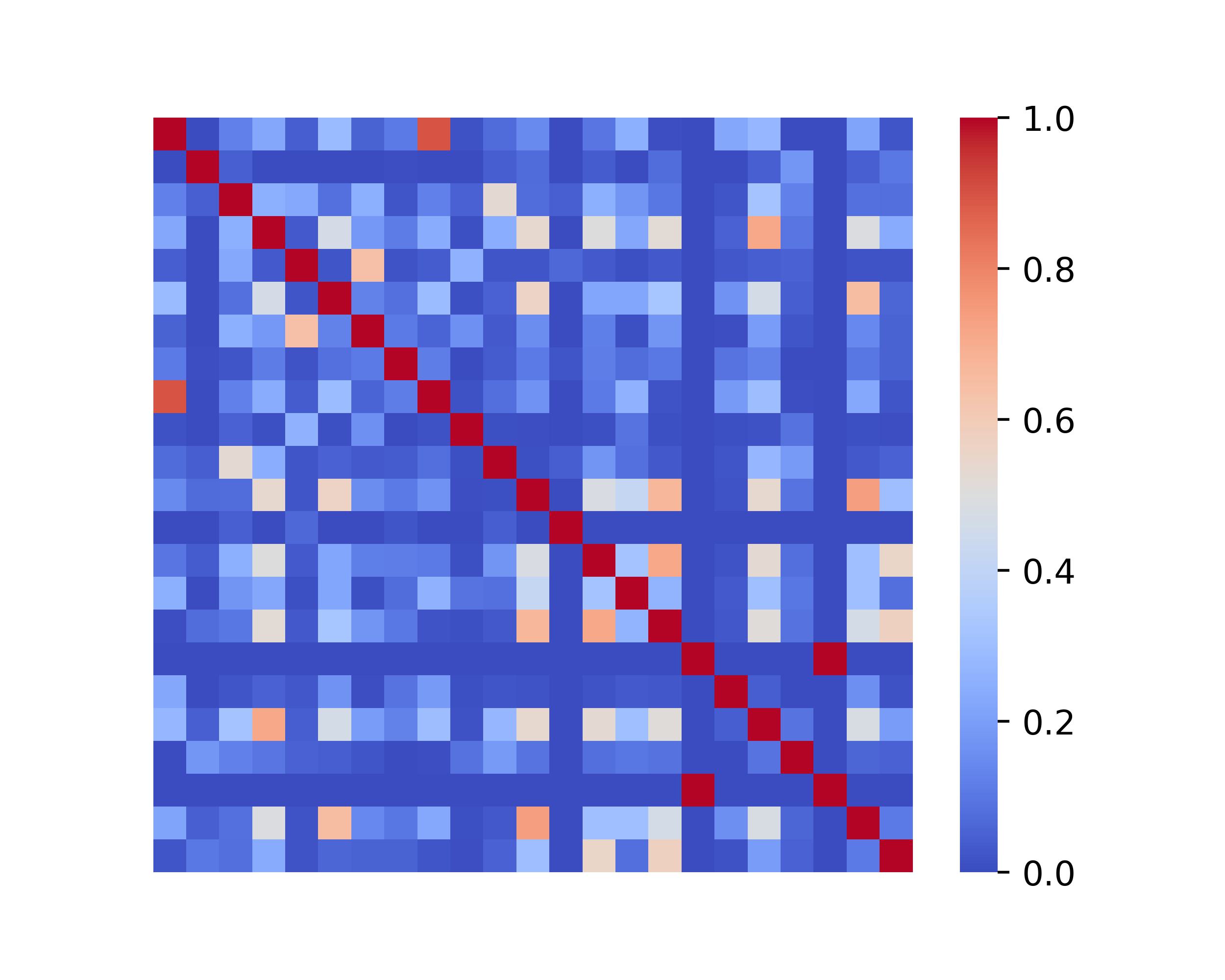}
\label{labelcor2}}
\subfigure[difference]{\includegraphics[width=0.14\textwidth]{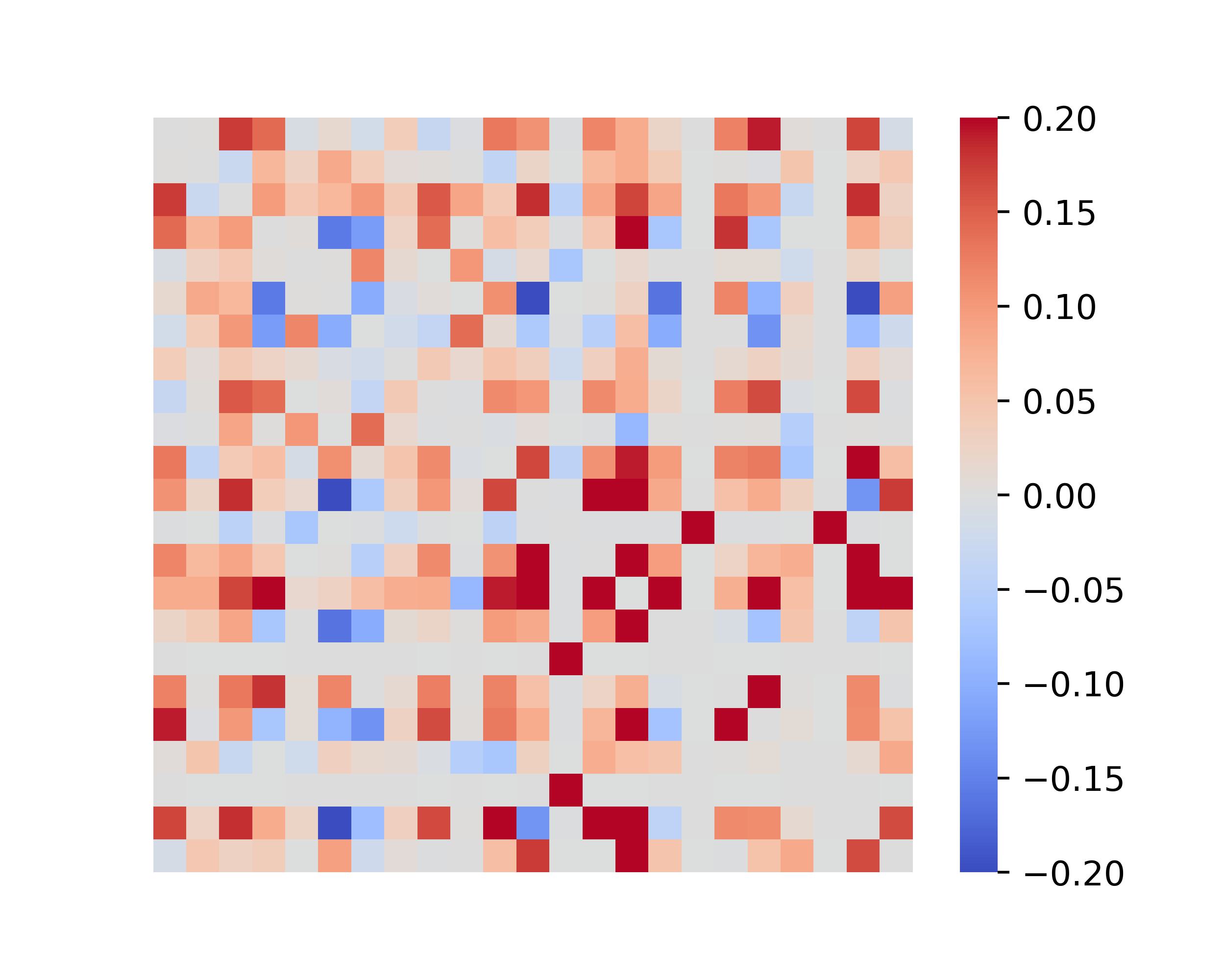}
\label{labelcordif}}
\caption{Visualization of label correlation matrix $\mathbf{C}$ change in \textit{CLIF} with Corel5k dataset.}
\end{figure}

The concept of uncertainty-based label correlation drift explores how the relationships between labels evolve during model training. Figures \ref{labelcor1} and \ref{labelcor2} illustrate the label correlation matrices at the 30th and 70th epochs, showing how the model's perception of label relationships, based on uncertainty, evolves during training. The differences highlighted in Figure \ref{labelcordif} reveal that these correlations change dynamically, suggesting that as the model learns, it adjusts its understanding of these uncertainty-based label relationships.

\section{Conclusion}
This paper proposes an uncertainty-based multi-label batch selection method, filling the gap in uncertainty-based multi-label batch selection. 
We introduce two key components to better adapt to the characteristics of multi-label data. First, for each label, we consider both the confidence of the current prediction and the changes between consecutive predictions within a sliding window, allowing for a more accurate assessment of label uncertainty. Second, we leverage dynamic uncertainty-based label correlations to comprehensively evaluate each sample's uncertainty, prioritizing samples that exhibit synergistic uncertainty across multiple labels during training. Experimental results show that our method greatly enhances model performance, speeds up convergence, and outperforms five other methods across diverse datasets and deep multi-label learning models.

In future work, temperature scaling can be explored to improve the entropy-based uncertainty assessment in deep learning models, and theoretical uncertainty methods based on Bayesian averaging, such as Monte Carlo dropout, can be considered.

\section*{Acknowledgements}
This work was supported by the National Natural Science Foundation of China (62302074) and the Science and Technology Research Program of Chongqing Municipal Education Commission (KJQN202300631).

\newpage


\section{Appendix A: Computation of Joint Probability}

Joint Probability \( p(u^\tau_{\cdot a},u^\tau_{\cdot b}) \):
The joint probability \( p(u^\tau_{\cdot a}, u^\tau_{\cdot b}) \) is calculated by counting the number of samples where the values \( u_{ij} \) from column \( \mathbf{u}_{\cdot a} \) fall into bin \( u^\tau_{\cdot a} \) and the values \( u_{ij} \) from column \( \mathbf{u}_{\cdot b} \) fall into bin \( u^\tau_{\cdot b} \), divided by the total number of instances \( n \) . Formally, it can be expressed as:
\begin{equation}
p(u^\tau_{\cdot a}, u^\tau_{\cdot b}) = \frac{\sum_{i=1}^{n} \mathbb{1}(u_{i, a} \in u^\tau_{\cdot a} \land u_{i, b} \in u^\tau_{\cdot b})}{n},
\end{equation}
where \( \mathbb{1}(\cdot) \) is an indicator function that returns 1 if the condition inside is true and 0 otherwise. The denominator \( n \) is the total number of instances.

Marginal Probability \( p(u^\tau_{\cdot a}) \):
The marginal probability \( p(u^\tau_{\cdot a}) \) is calculated by counting the number of instances where the values \( u_{ij} \) from column \( \mathbf{u}_{\cdot a} \) fall into bin \( u^\tau_{\cdot a} \), divided by the total number of instances \( n \) in the dataset. Formally, it can be expressed as:
\begin{equation}
p(u^\tau_{\cdot a}) = \frac{\sum_{i=1}^{n} \mathbb{1}(u_{i, a} \in u^\tau_{\cdot a})}{n}
\end{equation}
This equation provides the proportion of instances falling into the bin \( u^\tau_{\cdot a} \) for label \( l_a \). These probabilities are used to compute the mutual information \( C_{ab} \) as described in the provided equation.

\section{Appendix B: Convergency Analysis}
The index $Q(z) = \lfloor (1-z) /\Delta\rfloor$ of each instance is bounded by:
\begin{equation}
0 \leq Q(z) = \lfloor (1-z) /\Delta\rfloor \leq  n
\end{equation}
Then, the lower bound of $P(x_i) \supseteq  P(x_i|\mathcal{D}, w_i, n, s_t)$ is formulated by
\begin{equation}
0 < \frac{1}{\sum_{i=1}^{n} \frac{1}{\exp \left(\log(s_e)/{n} \right)^{Q(x)}}} \leq P(x_i) \supseteq  P(x_i|\mathcal{D}, w_i, n, s_t).
\end{equation}
Thus, the sampling distribution of multi-label adaptive selection is strictly positive.
To ensure a gradient estimate is unbiased, the expected value of the gradient estimate under the sampling distribution must equal the true gradient calculated across the entire dataset. Mathematically, this condition can be expressed as:
\begin{equation}
E_{ P(x_i|\mathcal{D}, w_i, n, s_t)}[\nabla L] = \nabla L_{\text{true}},
\end{equation}
where \(\nabla L\) is the gradient of the loss with respect to the model parameters, \(\nabla L_{\text{true}}\) is the true loss computed over the entire dataset, and \(E_{ P(x_i|\mathcal{D}, w_i, n, s_t)}[\cdot]\) denotes the expectation over the sampling distribution. Further, we can assume that if every sample in the training set has a probability of being selected, then the gradient is unbiased. 

Let us consider the gradient estimate $\tilde{G}$ used in combination with the Adam optimizer. Adam, an algorithm for first-order gradient-based optimization of stochastic objective functions, computes adaptive learning rates for each parameter. In its essence, Adam maintains two moving averages for each parameter: one for gradients ($m_t$) and one for the square of gradients ($v_t$). These moving averages are estimates of the first moment (the mean) and the second moment (the uncentered variance) of the gradients. The unbiasedness of the gradient estimate $\tilde{G}$, when combined with Adam, can be articulated by analyzing the correction step applied to $m_t$ and $v_t$. Specifically, the bias-corrected first and second moment estimates are given by:
\begin{align}
\hat{m}_t &= \frac{m_t}{1 - \beta_1^t}, \\
\hat{v}_t &= \frac{v_t}{1 - \beta_2^t},
\end{align}
where $\beta_1$ and $\beta_2$ are the exponential decay rates for the moment estimates, and $t$ denotes the timestep. The correction factors $(1 - \beta_1^t)$ and $(1 - \beta_2^t)$ counteract the bias towards zero in the initial time steps, ensuring that $\hat{m}_t$ and $\hat{v}_t$ are unbiased estimates of the first and second moments. Therefore, given an unbiased gradient estimate $\tilde{G}$, the application of Adam's bias correction guarantees that the adjusted gradients remain unbiased. 

\section{Appendix C: Comparison Methods}
\begin{itemize}
\item \textit{Random batch selection with shuffle (Base):} The default algorithm that shuffles the training data each epoch, selecting each batch randomly without replacement. [Shuffle: True].
\item \textit{Balanced batch selection (Balance) \cite{mllbalance}:} Ensures the proportion of labels in each batch matches the original training set using a stratified sampling approach. [Parameter: none].
\item \textit{Active Bias \cite{activebias}:} The uncertainty of the sample is calculated by adding the uncertainty of each label, which is based on the standard deviation of all historical period predictions. Samples with higher uncertainty are assigned higher weights. [Parameter: warm up epoch ($\gamma$): 5].
\item \textit{Recency Bias \cite{RecencyBias}:} The uncertainty of the sample is calculated by adding the uncertainty of each label, which is obtained based on the entropy of the sliding window binary prediction. Samples with greater uncertainty are given higher weights. [Parameter: Window size ($T$): 5, warm up epoch ($\gamma$): 5].
\item \textit{Hard and Imbalance batch selection (Hard) \cite{mllour}:} Prioritizes samples with higher loss and fewer associated labels, giving them greater weight for model training. [Parameter: selection pressure ($s_e$):  $\{2,8,16,64\}$, warm up epoch ($\gamma$): 5].
\end{itemize}

\section{Appendix D: Supplementary Results}
Tables \ref{ta:RankLoss} and \ref{ta:HammingLoss} present the experimental results of Ranking Loss and Hamming Loss on three classifiers, respectively.

The results presented in Table \ref{tab:Wilcoxon}, derived from the Wilcoxon signed-ranks test at a 0.05 significance level, conclusively indicate that our batch selection outperforms other batch selections with statistical significance.

\begin{table*}[t]
\centering
\caption{The Ranking Loss results of batch selection methods under different models, where the best results are highlighted by \textbf{boldface}.}  
\label{ta:RankLoss}
\resizebox{\textwidth}{!}{
\begin{tabular}{ccccccc|cccccc|cccccc}
\toprule
\multirow{2}{*}{Dataset} & \multicolumn{6}{c}{MPVAE} & \multicolumn{6}{c}{CLIF} & \multicolumn{6}{c}{DELA} \\ \cmidrule(lr){2-7} \cmidrule(lr){8-13} \cmidrule(lr){14-19}
& \textit{Random} & \textit{Balance} & \textit{Active} & \textit{Recent} & \textit{Hard} & \textit{Ours} & \textit{Random} & \textit{Balance} & \textit{Active} & \textit{Recent} & \textit{Hard} & \textit{Ours} & \textit{Random} & \textit{Balance} & \textit{Active} & \textit{Recent} & \textit{Hard} & \textit{Ours} \\ \midrule
scene & 0.0968(6) & 0.0898(4) & 0.0869(2) & 0.0869(2) & 0.0942(5) & \textbf{0.0864(1)} 
& 0.0891(6) & 0.0854(3) & 0.0861(5) & 0.0857(4) & 0.0850(2) & \textbf{0.0814(1)} 
& 0.0898(6) & 0.0846(4) & 0.0846(4) & 0.0836(3) & 0.0834(2) & \textbf{0.0825(1)} \\

yeast & 0.2011(5) & 0.2045(6) & 0.2003(4) & 0.1989(3) & 0.1981(2) & \textbf{0.1972(1)} 
& 0.1995(6) & 0.1970(5) & 0.1965(4) & 0.1952(3) & 0.1930(2) & \textbf{0.1923(1)} 
& 0.1975(6) & 0.1972(5) & 0.1964(4) & 0.1926(2) & 0.1933(3) & \textbf{0.1919(1)} \\

Corel5k & 0.0433(4) & 0.0405(3) & 0.0466(6) & 0.0442(5) & 0.0393(2) & \textbf{0.0371(1)} 
& 0.0231(3) & 0.0246(6) & 0.0232(4) & 0.0228(2) & 0.0240(5) & \textbf{0.0214(1)} 
& \textbf{0.0216(1)} & 0.0225(4) & 0.0225(4) & 0.0231(6) & 0.0223(3) & 0.0220(2) \\

rcv1subset1& 0.0399(6) & \textbf{0.0342(1)} & 0.0376(4) & 0.0372(3) & 0.0380(5) &0.0358(2) 
& 0.0355(6) & 0.0339(5) & 0.0316(3) & 0.0310(2) & 0.0338(4) & \textbf{0.0295(1)} 
& 0.0362(6) & 0.0360(5) & 0.0355(2) & 0.0358(4) & 0.0356(3) & \textbf{0.0343(1)} \\

rcv1subset2& 0.0369(6) & 0.0346(4) & 0.0320(3) & 0.0314(2) & 0.0346(4) & \textbf{0.0299(1)} 
& 0.0302(6) & 0.0291(2) & 0.0293(4) & 0.0291(2) & 0.0299(5) & \textbf{0.0289(1)} 
& 0.0312(6) & 0.0309(5) & 0.0306(4) & 0.0300(2) & 0.0303(3) & \textbf{0.0297(1)} \\

rcv1subset3& 0.0362(6) & 0.0335(4) & 0.0332(3) & 0.0325(2) & 0.0344(5) & \textbf{0.0316(1)} 
& 0.0341(6) & 0.0304(5) & 0.0294(4) & 0.0289(3) & 0.0282(2) & \textbf{0.0274(1)} 
& 0.0305(6) & 0.0298(4) & 0.0293(3) & 0.0291(2) & 0.0300(5) & \textbf{0.0286(1)} \\

yahoo-Business1 & 0.0452(5) & 0.0485(6) & 0.0438(3) & 0.0427(2) & 0.0446(4) & \textbf{0.0416(1)} 
& 0.0165(3) &0.0166(4)  & 0.0172(6) & 0.0168(5) & 0.0163(2) & \textbf{0.0160(1)} 
& 0.0173(4) & 0.0181(6) & 0.0171(3) & 0.0179(5) & 0.0167(2) & \textbf{0.0156(1)} \\

yahoo-Arts1 & 0.0528(3) & 0.0573(6) & 0.0551(5) & 0.0526(2) & 0.0538(4) & \textbf{0.0512(1)} 
& 0.0556(5) & 0.0553(4) & 0.0548(2) & \textbf{0.0540(1)} & 0.0561(6) & 0.0549(3) 
& 0.0547(6) & 0.0544(4) & 0.0545(5) & 0.0535(2) & 0.0537(3) & \textbf{0.0526(1)} \\

bibtex& 0.0194(3) & 0.0210(4) & 0.0242(6) & 0.0214(5) & 0.0147(2) & \textbf{0.0134(1)} 
& 0.0133(3) & 0.0151(6) & 0.0134(4) & 0.0139(5) & 0.0132(2) & \textbf{0.0126(1)} 
& \textbf{0.0133(1)} & 0.0159(6) & 0.0154(5) & 0.0148(4) & 0.0137(2) & 0.0139(3) \\

tmc2007& 0.0700(3) & 0.0781(6) & 0.0759(5) & 0.0710(4) & 0.0685(2) & \textbf{0.0634(1)} 
& 0.0700(5) & 0.0736(6) & 0.0692(3) & 0.0688(2) & 0.0697(4) & \textbf{0.0682(1)} 
& 0.0673(3) & 0.0688(6) & 0.0674(4) & 0.0682(5) & 0.0667(2) & \textbf{0.0658(1)} \\

enron& 0.0901(6) & 0.0832(4) & 0.0767(3) & \textbf{0.0755(1)} & 0.0840(5) & 0.0762(2) 
& 0.0758(5) & 0.0764(6) & 0.0752(4) & 0.0744(2) & 0.0745(3) & \textbf{0.0729(1)} 
& 0.0735(4) & 0.0746(6) & 0.0741(5) & 0.0718(2) & 0.0722(3) & \textbf{0.0715(1)} \\

cal500& 0.2416(2) & 0.2487(6) & 0.2428(5) & 0.2424(4) & 0.2417(3) & \textbf{0.2406(1)} 
& 0.1900(3) & 0.1903(4) & 0.1924(6) & 0.1912(5) & 0.1895(2) & \textbf{0.1893(1)} 
& 0.1894(2) & 0.1907(5) & 0.1898(4) & 0.1914(6) & 0.1894(2) & \textbf{0.1878(1)} \\

LLOG-F& 0.2075(6) & 0.1980(5) & 0.1966(3) & \textbf{0.1916(1)} & 0.1976(4) & 0.1935(2) 
& 0.1892(6) & 0.1694(5) & 0.1688(3) & \textbf{0.1675(1)} & 0.1691(4) & 0.1682(2)
& 0.1603(5) & 0.1591(3) & 0.1596(4) & 0.1589(2) & 0.1604(6) & \textbf{0.1582(1)} \\ \midrule
Avg & 4.69 & 4.54 & 4.00 & 2.77 & 3.62 & \textbf{1.23} & 4.85 & 4.69 & 4.00 & 2.85 & 3.31 & \textbf{1.23} & 4.31 & 4.85 & 3.92 & 3.46 & 3.00 & \textbf{1.23} \\
\bottomrule
\end{tabular}}
\end{table*}

\begin{table*}[h]
\centering
\caption{The Hamming Loss results of batch selection methods under different models, where the best results are highlighted by \textbf{boldface}.}  
\label{ta:HammingLoss}
\resizebox{\textwidth}{!}{
\begin{tabular}{ccccccc|cccccc|cccccc}
\toprule
\multirow{2}{*}{Dataset} & \multicolumn{6}{c}{MPVAE} & \multicolumn{6}{c}{CLIF} & \multicolumn{6}{c}{DELA} \\ \cmidrule(lr){2-7} \cmidrule(lr){8-13} \cmidrule(lr){14-19}
& \textit{Random} & \textit{Balance} & \textit{Active} & \textit{Recent} & \textit{Hard} & \textit{Ours} & \textit{Random} & \textit{Balance} & \textit{Active} & \textit{Recent} & \textit{Hard} & \textit{Ours} & \textit{Random} & \textit{Balance} & \textit{Active} & \textit{Recent} & \textit{Hard} & \textit{Ours} \\ \midrule
scene& 0.0920(6) & 0.0831(2) & 0.0889(4) & 0.0840(3) & 0.0892(5) & \textbf{0.0826(1)}
& 0.0722(5) & 0.0695(2) & 0.0699(4) & \textbf{0.0684(1)} & 0.0726(6) & 0.0698(3)
& 0.0725(6) & 0.0693(5) & \textbf{0.0653(1)} & 0.0662(3) & 0.0676(4) & 0.0657(2) \\
yeast& 0.1719(3) & 0.1736(6) & 0.1720(4) & \textbf{0.1704(1)} & 0.1723(5) & 0.1712(2)
& 0.1677(5) & 0.1696(6) & 0.1673(4) & 0.1657(3) & 0.1631(2) & \textbf{0.1626(1)}
& 0.1679(5) & 0.1696(6) & 0.1643(3) & 0.1628(2) & \textbf{0.1607(1)} & 0.1658(4) \\
Corel5k & 0.2822(2) & 0.2865(6) & 0.2844(5) & 0.2826(3) & 0.2832(4) & \textbf{0.2814(1)}
& 0.1550(3) & 0.1580(6) & 0.1564(4) & 0.1574(5) & 0.1532(2) & \textbf{0.1525(1)}
& 0.1577(2) & 0.1696(5) & 0.1596(3) & 0.1600(4) & 0.1762(6) & \textbf{0.1508(1)} \\
rcv1subset1 & 0.0930(3) & 0.0967(6) & 0.0945(5) & 0.0936(4) & 0.0917(2) & \textbf{0.0912(1)}
& 0.0576(5) & 0.0578(6) & 0.0570(3) & 0.0566(2) & 0.0571(4) & \textbf{0.0564(1)}
& 0.0617(6) & 0.0562(2) & 0.0582(5) & 0.0579(4) & 0.0576(3) & \textbf{0.0532(1)} \\
rcv1subset2 & 0.1014(4) & 0.1006(3) & 0.1022(5) & 0.0986(2) & 0.1067(6) & \textbf{0.0943(1)}
& 0.0540(4) & 0.0553(6) & 0.0542(5) & 0.0539(3) & 0.0538(2) & \textbf{0.0518(1)}
& 0.0611(5) & 0.0602(2) & 0.0611(5) & 0.0603(3) & 0.0610(4) & \textbf{0.0576(1)} \\
rcv1subset3 & 0.1187(4) & 0.1162(2) & 0.1194(6) & 0.1172(3) & 0.1187(4) & \textbf{0.1126(1)}
& 0.0569(3) & 0.0584(6) & 0.0575(4) & 0.0564(2) & 0.0581(5) & \textbf{0.0556(1)}
& 0.0606(5) & 0.0609(6) & 0.0596(2) & 0.0600(3) & 0.0602(4) & \textbf{0.0589(1)} \\
yahoo-Business1& 0.0373(2) & 0.0418(6) & 0.0406(5) & 0.0386(4) & 0.0379(3) & \textbf{0.0368(1)}
& 0.0358(4) & 0.0408(6) & 0.0384(5) & 0.0340(2) & 0.0343(3) & \textbf{0.0337(1)}
& 0.0430(6) & 0.0392(5) & 0.0380(4) & 0.0377(3) & 0.0374(2) & \textbf{0.0364(1)} \\
yahoo-Arts1 & 0.1188(6) & 0.1170(5) & 0.1145(4) & 0.1120(2) & 0.1129(3) & \textbf{0.1106(1)}
& 0.1227(4) & 0.1268(6) & 0.1236(5) & 0.1214(3) & 0.1202(2) & \textbf{0.1189(1)}
& 0.1293(5) & 0.1308(6) & 0.1263(4) & 0.1235(2) & \textbf{0.1207(1)} & 0.1238(3) \\
bibtex & 0.1218(3) & 0.1390(6) & 0.1248(5) & 0.1225(4) & 0.1216(2) & \textbf{0.1202(1)}
& 0.0742(2) & 0.0765(6) & 0.0759(5) & 0.0753(4) & 0.0746(3) & \textbf{0.0737(1)}
& 0.0884(4) & 0.1009(6) & 0.0876(3) & 0.0899(5) & 0.0841(2) & \textbf{0.0822(1)} \\
tmc2007 & 0.0785(4) & 0.0794(6) & 0.0789(5) & 0.0770(3) & 0.0764(2) & \textbf{0.0752(1)} 
& 0.0524(3) & 0.0574(6) & 0.0563(5) & 0.0546(4) & 0.0517(2) & \textbf{0.0503(1)} 
& 0.0552(4) & 0.0566(6) & 0.0540(3) & \textbf{0.0533(1)} & 0.0552(4) & 0.0537(2) \\
enron & 0.1610(6) & 0.1520(4) & 0.1562(5) & 0.1508(2) & \textbf{0.1496(1)} & 0.1516(3)
& 0.1154(6) & 0.1137(5) & 0.1125(4) & 0.1108(2) & 0.1112(3) & \textbf{0.1106(1)} 
& 0.1232(6) & 0.1202(3) & 0.1212(5) & \textbf{0.1194(1)}  & 0.1198(2) & 0.1206(4) \\
cal500& 0.2410(2) & 0.2421(4) & 0.2477(6) & 0.2462(5) & 0.2413(3) & \textbf{0.2406(1)} 
& 0.2312(4) & 0.2300(3) & 0.2314(5) & 0.2335(6) & 0.2273(2) & \textbf{0.2268(1)} 
& 0.2210(2) & 0.2257(5) & 0.2253(4) & 0.2228(3) & 0.2259(6) & \textbf{0.2181(1)} \\
LLOG-F& 0.2011(6) & 0.1981(5) & 0.1937(3) & \textbf{0.1914(1)} & 0.1961(4) & 0.1932(2) 
& 0.1865(6) & 0.1856(5) & 0.1847(3) & \textbf{0.1832(1)} & 0.1849(4) & 0.1840(2) 
& 0.2169(6) & 0.2152(5) & 0.2147(4) & 0.2135(2) & 0.2138(3) & \textbf{0.2125(1)} \\ \midrule
Avg & 3.92 & 4.69 & 4.77 & 2.85 & 3.38 & \textbf{1.31}
& 4.15 & 5.31 & 4.31 & 2.92 & 3.08 & \textbf{1.23}
& 4.77 & 4.77 & 3.54 & 2.77 & 3.23 & \textbf{1.77} \\\bottomrule
\end{tabular}}
\end{table*}

\begin{table*}[!h]
\centering
\caption{P-values of the Wilcoxon Signed-Ranks Test for adaptive against our method in terms of each evaluation metric.}
\label{tab:Wilcoxon}
\resizebox{\textwidth}{!}{
\begin{tabular}{cccccc|ccccc|ccccc}
\toprule
\multirow{2}{*}{Metric} & \multicolumn{5}{c}{MPVAE} & \multicolumn{5}{c}{CLIF} & \multicolumn{5}{c}{DELA} \\ \cmidrule(lr){2-6} \cmidrule(lr){7-11} \cmidrule(lr){12-16}
& \textit{Random} & \textit{Balance} & \textit{Active} & \textit{Recent} & \textit{Hard} & \textit{Random} & \textit{Balance} & \textit{Active} & \textit{Recent} & \textit{Hard}  & \textit{Random} & \textit{Balance} & \textit{Active} & \textit{Recent} & \textit{Hard} \\ \midrule
Macro-AUC & 0.0002 & 0.0002 & 0.0002 & 0.0002 & 0.0007 & 0.0002 & 0.0002 & 0.0007 & 0.0134 & 0.0007 & 0.0005 & 0.0002 & 0.0002 & 0.0942 & 0.0017 \\
Ranking Loss &0.0002 & 0.0005 & 0.0002 & 0.0171 & 0.0002 & 0.0002 & 0.0002 & 0.0005 & 0.0081 & 0.0002 & 0.0012 & 0.0002 & 0.0002 & 0.0002 & 0.0005 \\
Hamming Loss &0.0002 & 0.0002 & 0.0002 & 0.0061 & 0.0061 & 0.0002 & 0.0005 & 0.0002 & 0.0017 & 0.0002 & 0.0002  & 0.0002 & 0.0005 & 0.0061 & 0.0061 \\ \bottomrule
\end{tabular}
}
\end{table*}

\section{Appendix E: Computational Complexity Analysis}

\begin{figure}[!h]
    \centering
    \includegraphics[width=0.7\linewidth]{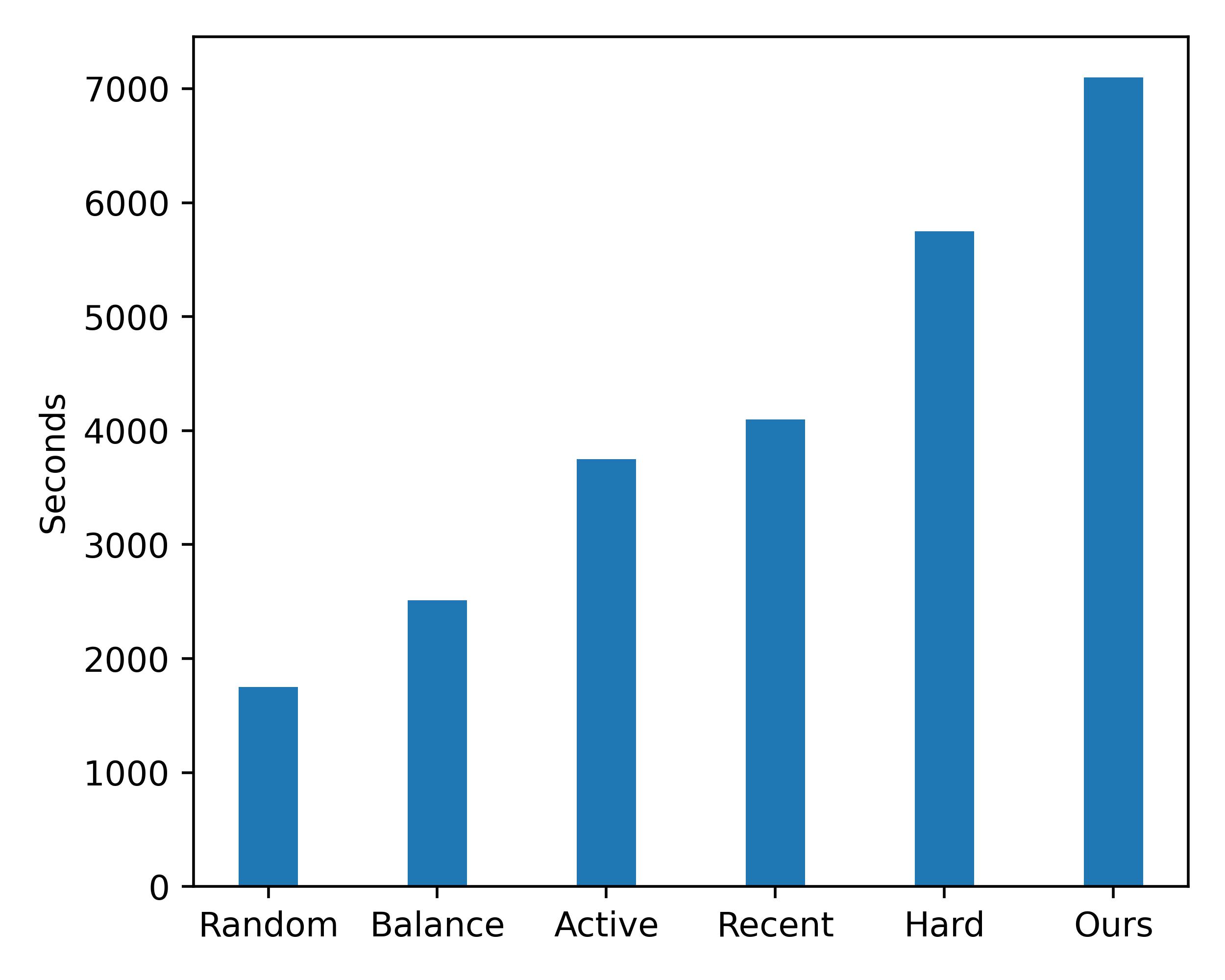}
    \caption{The time required for the batch selection method using \textit{CLIF} to complete 100 epochs on the yahoo-Arts dataset.}
    \label{fig:time}
\end{figure}

The computational complexities of comparing and the proposed methods are listed in Table \ref{ta:computation}.
With all methods training for the same number of epochs, the last two methods are more time-consuming, as confirmed by our newly obtained training time comparison results shown in Figure \ref{fig:time}.

\begin{table}[h]
\centering
\caption{Computational complexity of different batch selection method}
\label{ta:computation}
\begin{tabular}{|c|c|}
\hline
Method & \textbf{Computational Complexity} \\
\hline
Random& \( O(1) \) \\
\hline
Balance & \( O(nq) \) \\
\hline
Active and Recent & \( O(nqT) \) \\
\hline
Hard & \( O(n + d + nqk) \) \\
\hline
Ours & \( O(nq(T + q)) \) \\
\hline
\end{tabular}
\end{table}

Nevertheless, our method converges faster than the baselines, meaning our method achieves the best validation result in fewer epochs. For example, for yahoo-Arts with the \textit{CLIF} model, Random achieves the best AUC at epoch 85, taking 1560 seconds, while our method reaches the same AUC result at epoch 32 using only 1280 seconds.

\section{Appendix F: Ablation Study}
In this section, we conduct several experiments to understand the importance of several key components in our method. A denotes the full method. 
Variant B only considers the uncertainty of current predictions ($\lambda_1 = 0$).  
Variant C only considers fluctuations of predictions within $T$ sliding windows ($\lambda_1 = 1$).
Variant D dose not use the uncertainty-based label correlations matrix $\mathbf{C}$.
\begin{figure}[!h]
\centering
\subfigure[bibtex]{\includegraphics[width=0.38\textwidth]{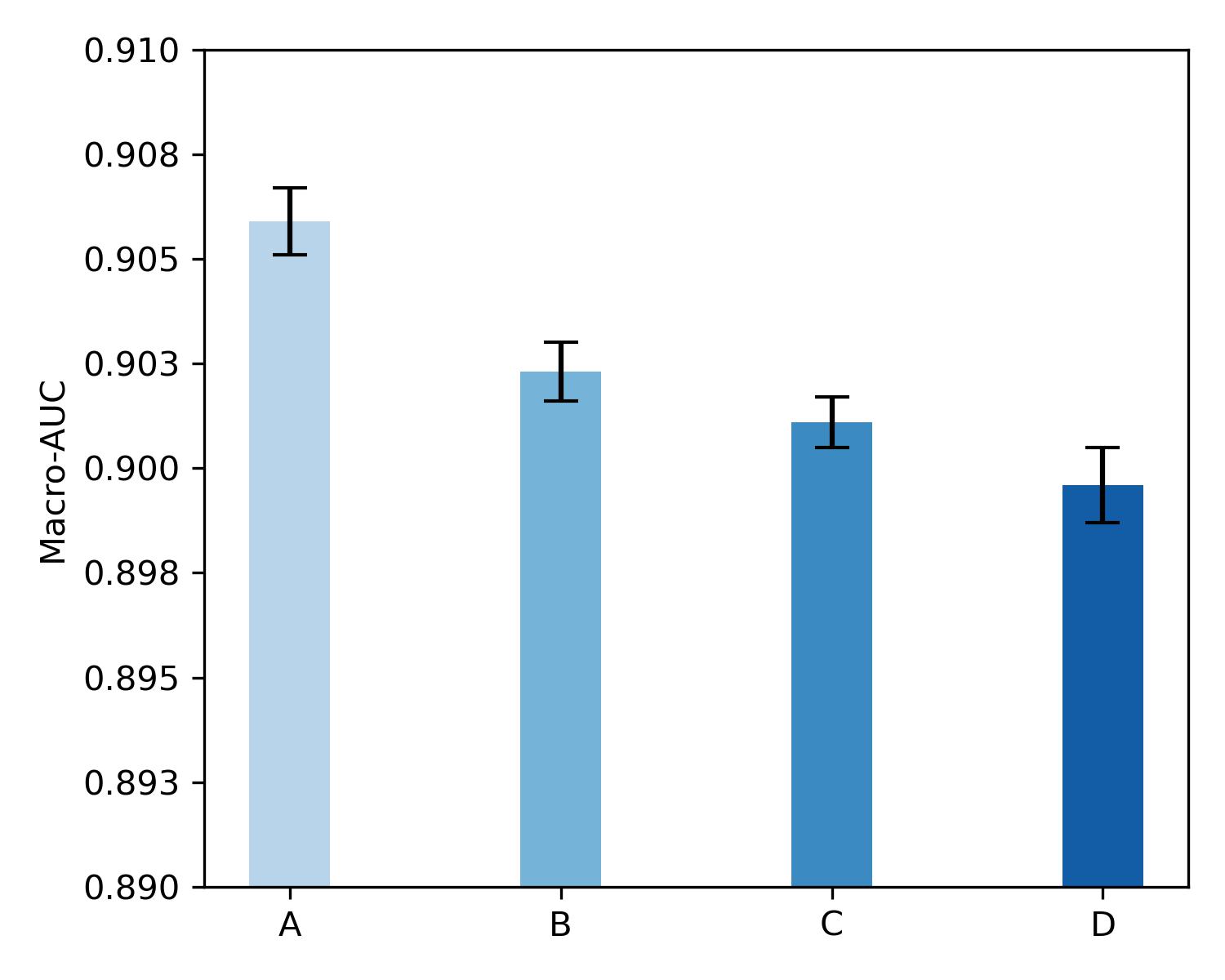}} 
\caption{Ablation experiment using \textit{CLIF}.}
\label{fig:ablation_results}
\end{figure}
Figure \ref{fig:ablation_results} presents the ablation study results for the bibtex dataset under \textit{CLIF}. 
The full method (A) achieves the highest Macro-AUC scores, illustrating its comprehensive effectiveness. In contrast, variant B, which only considers the uncertainty of current predictions, results in the lowest performance, highlighting the significance of temporal context. Variant C, utilizing uncertainty from sliding windows, shows improvement over B but still falls short compared to A. Lastly, variant D, which omits the uncertainty-based label correlations matrix, demonstrates a notable decrease in performance, underlining the critical role of integrating label correlations. These findings collectively emphasize that optimal performance is achieved by incorporating all components, including both current and historical uncertainties alongside the label correlation matrix.


Furthermore, we explore the combination of the Active \cite{activebias} and Recent \cite{RecencyBias} batch selection methods with the incorporation of the label correlation matrix \(\mathbf{C}\) (denoted as $Active_LC$ and $Recent_LC$, respectively).

\begin{figure}[!h]
    \centering
    \includegraphics[width=0.7\linewidth]{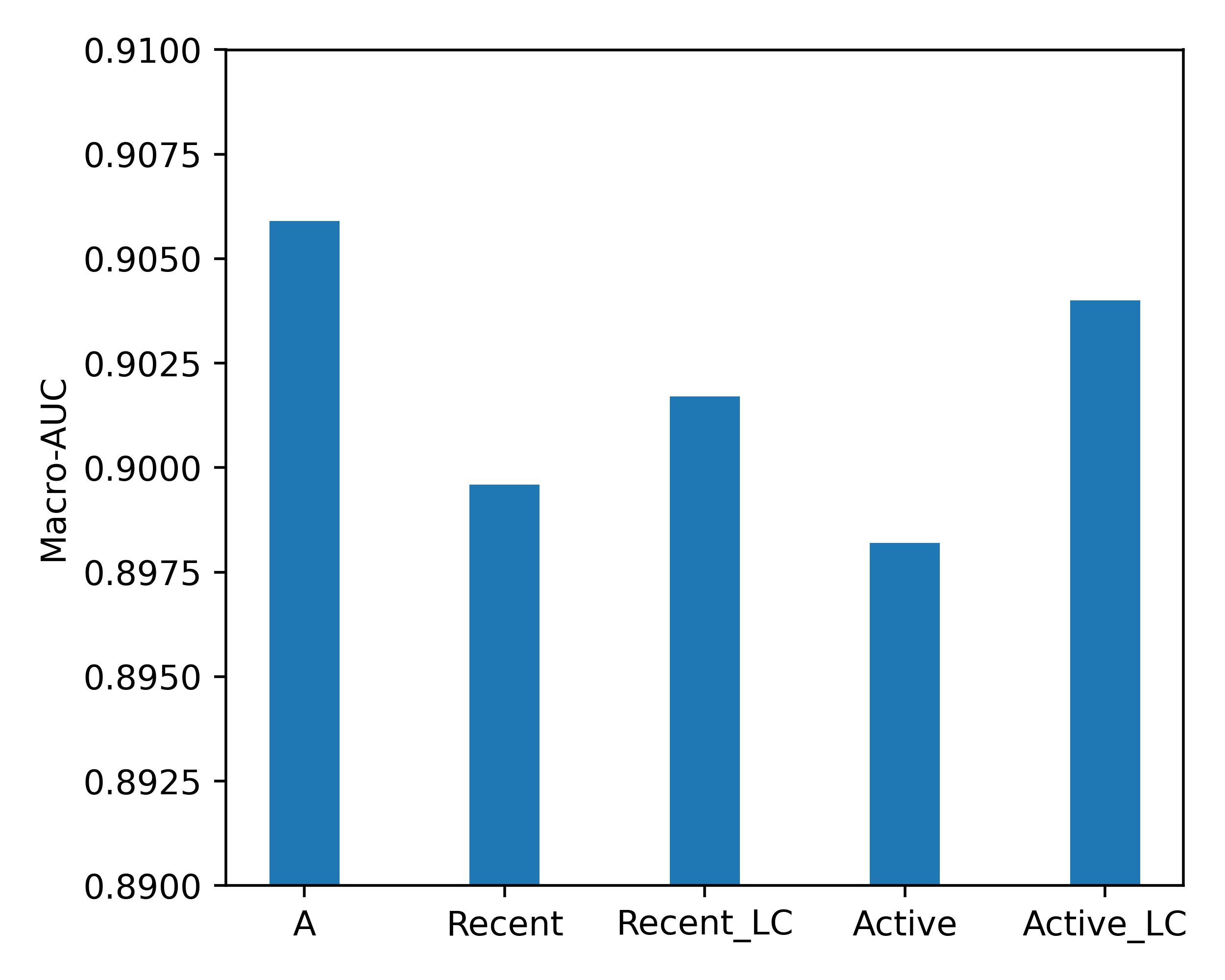}
    \caption{Ablation experiment using \textit{CLIF} on bibtex dataset.}
    \label{fig:ablation_result}
\end{figure} 
Figure \ref{fig:ablation_result} presents the result for the bibtex dataset under \textit{CLIF}. 
Variants $Recent_LC$ and $Active_LC$, which incorporate label correlations, outperform Recent and Active, respectively, demonstrating the effectiveness of the uncertainty-based label correlation matrix $\mathbf{C}$. 
However, they still fall short of the A due to their limited ability to assess uncertainty comprehensively.

\section{Appendix G: Parameter Analysis}

\begin{figure}[h]
\centering
\subfigure[Corel5k]{\includegraphics[width=0.2\textwidth]{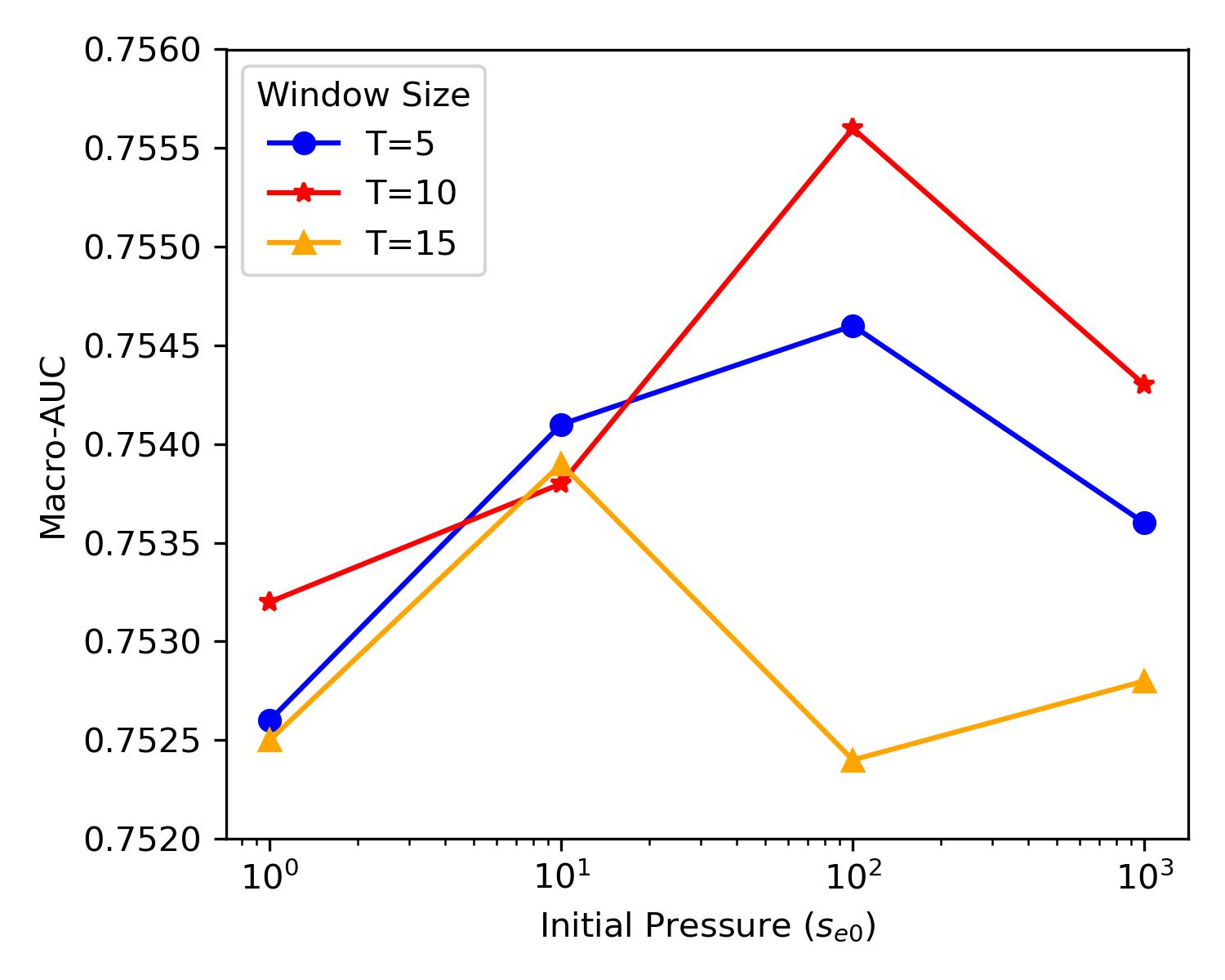}} 
\hspace{0.05\textwidth} 
\subfigure[bibtex]{\includegraphics[width=0.2\textwidth]{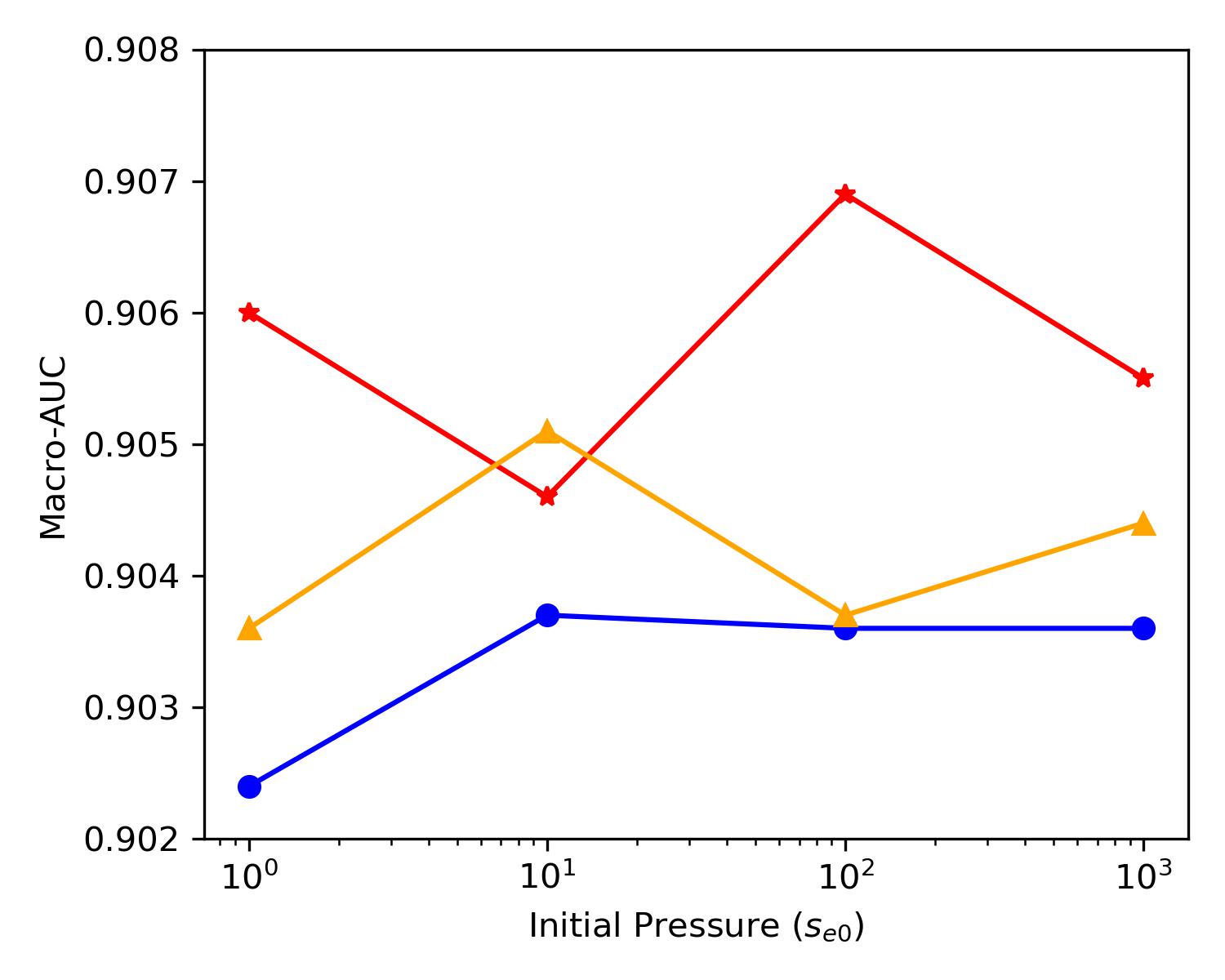}}
\caption{Grid search on Corel5k and bibtex datasets using \textit{CLIF}.}
\label{fig:Ablation_sliding_window}
\end{figure}

Figure \ref{fig:Ablation_sliding_window} shows the Macro-AUC obtained by grid search on two datasets using \textit{CLIF}. For the initial selection pressure $s_t$, a value of 100 typically yields the lowest testing error. For the window size $T$, a value of 5 consistently results in the lowest testing error. Similar trends are observed with other base classifiers and optimizer combinations. Therefore, in all experiments, we set $s_t$ 100 to 0 and $T$ to 5.

\begin{figure}[h]
\centering
\subfigure[Corel5k]{\includegraphics[width=0.2\textwidth]{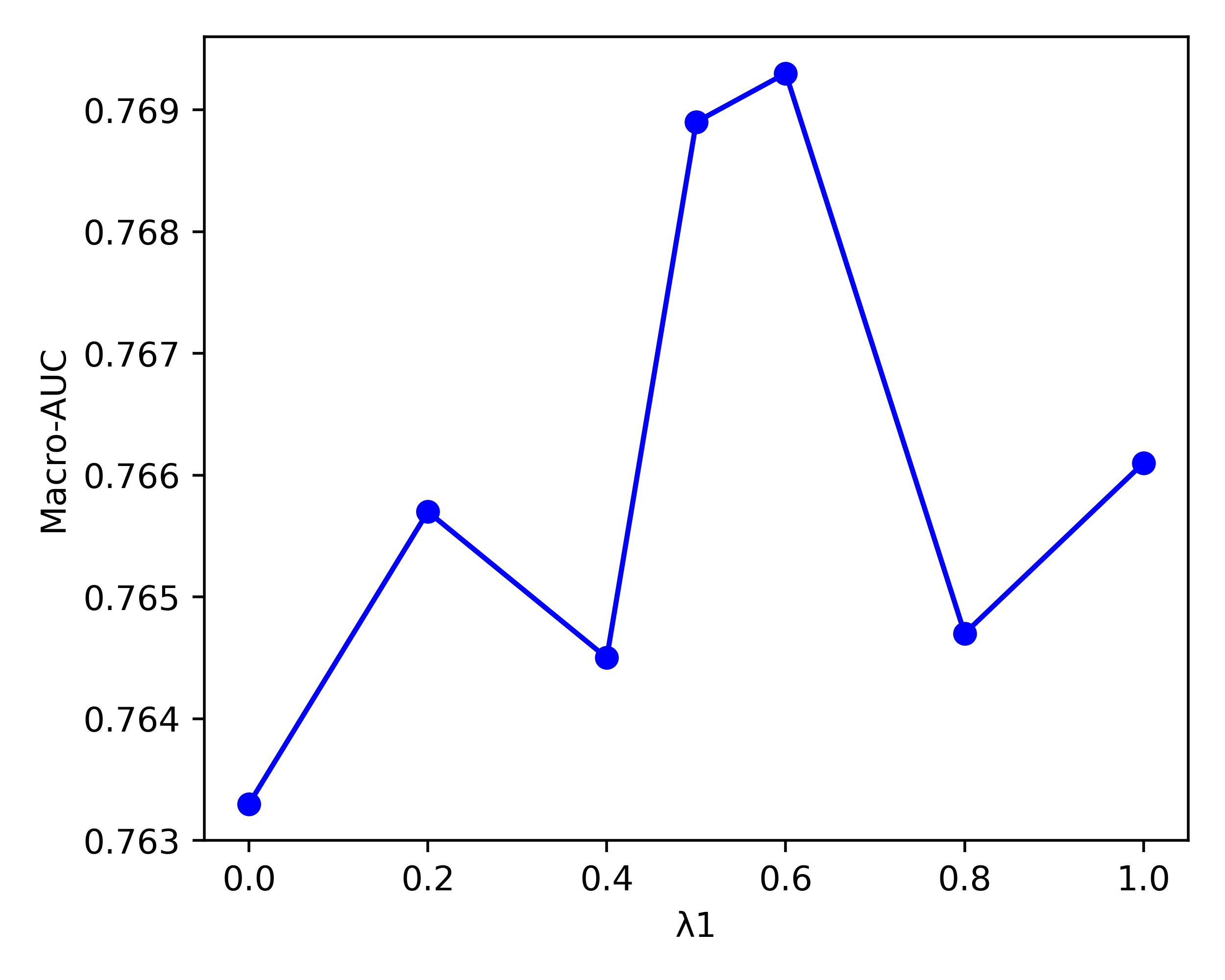}} 
\hspace{0.05\textwidth} 
\subfigure[bibtex]{\includegraphics[width=0.2\textwidth]{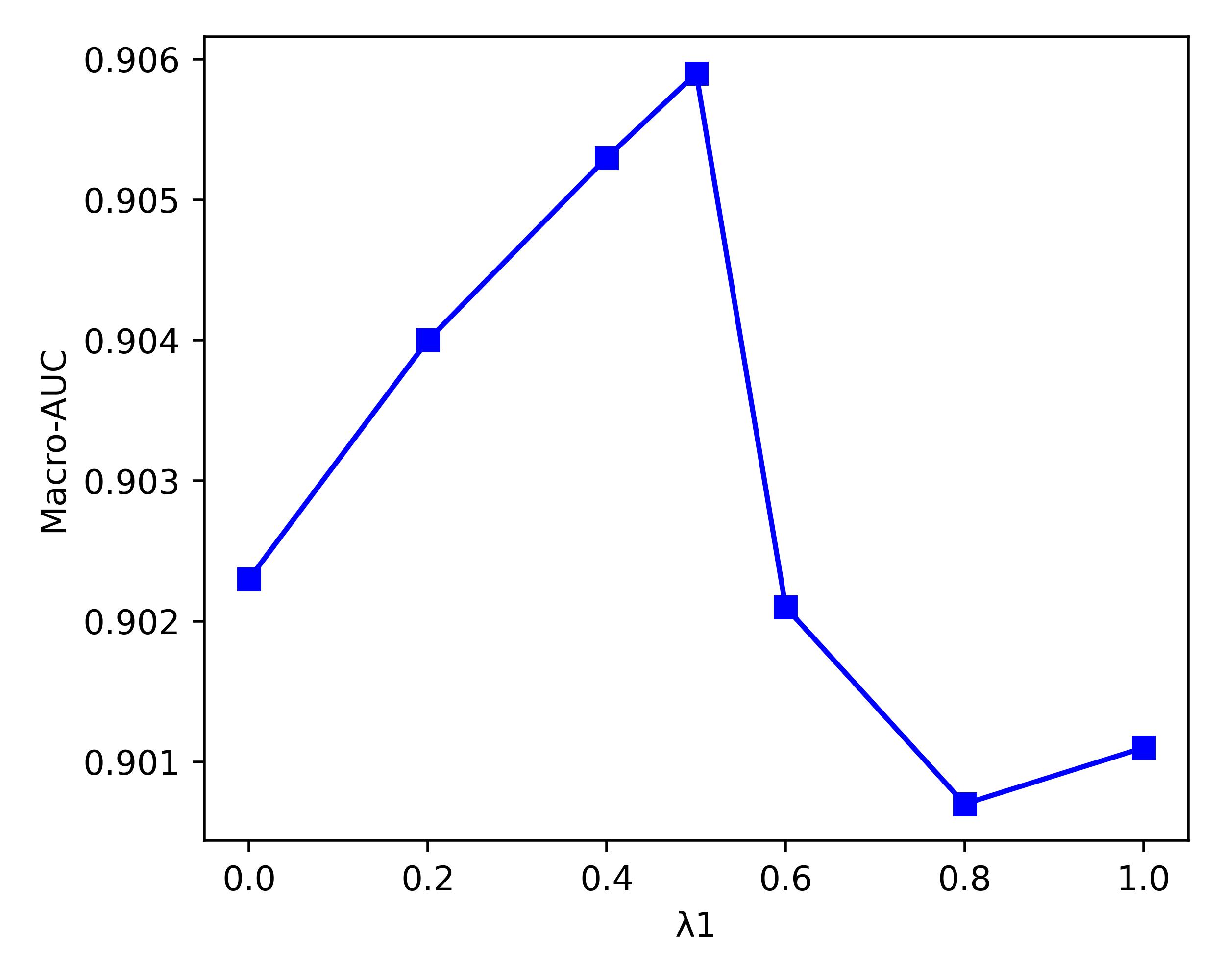}}
\caption{$\lambda_1$ sensitivity analysis on Corel5k and bibtex datasets using \textit{CLIF}.}
\label{fig:lambda_sensitive}
\end{figure}

As shown in Figure \ref{fig:lambda_sensitive}, we further conduct a sensitivity analysis on $\lambda_1$. The results indicate optimal performance within the range $[0.5–0.6]$, while extremely larger and lower values lead to lower accuracy due to neglecting one uncertainty factor. This implies that $[0.5-0.6]$ is a suggested range to search for the optimal $\lambda_1$ and obtain better results. 

\begin{thebibliography}{38}
\bibitem[{Hand, Castillo, and Chellappa(2018)}]{mllbalance}
Hand, E.; Castillo, C.; and Chellappa, R. 2018.
\newblock Doing the best we can with what we have: multi-label balancing with selective learning for attribute prediction.
\newblock In \emph{AAAI}, 6878–6885.
\end{thebibliography}
\end{document}